\documentclass{article}

\PassOptionsToPackage{dvipsnames}{xcolor}

\usepackage[preprint]{neurips_2026}

\usepackage[utf8]{inputenc} % allow utf-8 input
\usepackage[T1]{fontenc}    % use 8-bit T1 fonts
\usepackage{hyperref}       % hyperlinks
\usepackage{url}            % simple URL typesetting
\usepackage{booktabs}       % professional-quality tables
\usepackage{amsfonts}       % blackboard math symbols
\usepackage{nicefrac}       % compact symbols for 1/2, etc.
\usepackage{microtype}      % microtypography
\usepackage{xcolor}         % colors
\usepackage[table]{xcolor}

\usepackage{graphicx}
\usepackage{subcaption}
\usepackage{bm}
\usepackage[most]{tcolorbox}
\usepackage{amsmath}
\usepackage{amssymb}
\usepackage{mathtools}
\usepackage{amsthm}
\usepackage{multirow}
\usepackage{makecell}
\usepackage{pifont}
\usepackage[capitalize,noabbrev]{cleveref}
\usepackage{xspace}
\usepackage{algorithm}
\usepackage{algorithmic}
\usepackage{wrapfig}
\usepackage{etoc}
\usepackage{titletoc}
\usepackage{enumitem}
\usepackage{array}
\theoremstyle{plain}

\theoremstyle{definition}

\theoremstyle{remark}

\definecolor{checkgreen}{RGB}{76,175,80} % #4CAF50
\definecolor{lightblue}{rgb}{0.88,0.92,1}
\definecolor{resultgreen}{HTML}{689d6A}
\definecolor{resultred}{HTML}{FB4934}
\definecolor{resultcomparable}{HTML}{a89984}
\newcommand{\good}[1]{\textcolor{resultgreen}{#1}}
\newcommand{\bad}[1]{\textcolor{resultred}{#1}}
\newcommand{\comp}[1]{\textcolor{resultcomparable}{#1}}
\newcommand{\cmark}{\good{\ding{51}}}
\newcommand{\xmark}{\bad{\ding{55}}}

\usepackage{amsmath,amsfonts,bm,eqnarray}
\usepackage{cancel}
\usepackage{amsthm}
\usepackage{tikz}
\usetikzlibrary{tikzmark}

\newtheorem{innercustomgeneric}{\customgenericname}
\providecommand{\customgenericname}{}
\newcommand{\newcustomtheorem}[2]{%
  \newenvironment{#1}[1]
  {%
   \renewcommand\customgenericname{#2}%
   \renewcommand\theinnercustomgeneric{##1}%
   \innercustomgeneric
  }
  {\endinnercustomgeneric}
}

\newcustomtheorem{customThm}{Theorem}
\newcustomtheorem{customLemma}{Lemma}
\newcustomtheorem{customCor}{Corollary}
\newcustomtheorem{customProposition}{Proposition}

\def\eqref#1{equation~\ref{#1}}
\def\Eqref#1{Eqn.~\ref{#1}}

\def\1{\bm{1}}

\DeclareMathAlphabet{\mathsfit}{\encodingdefault}{\sfdefault}{m}{sl}
\SetMathAlphabet{\mathsfit}{bold}{\encodingdefault}{\sfdefault}{bx}{n}

\renewcommand{\tilde}{\widetilde}
\renewcommand{\hat}{\widehat}
\renewcommand{\frac}{\tfrac}

\renewcommand{\cite}{\citep}

\title{SNLP: Layer-Parallel Inference via Structured Newton Corrections}

\author{%
  \vspace{-4mm}\\
  \textbf{Ligong Han\textsuperscript{1,2,*,\raisebox{0.35ex}{$\dagger$}}~~~~Kai Xu\textsuperscript{1,2,*}~~~~Hao Wang\textsuperscript{1,2}~~~~Akash Srivastava\textsuperscript{2,3}}\\[0.5mm]
  \textsuperscript{1}Red Hat AI Innovation~~~~\textsuperscript{2}MIT-IBM Watson AI Lab~~~~\textsuperscript{3}Core AI, IBM\\[-0.2mm]
  {\small \textsuperscript{*}Equal contribution ~~~~\textsuperscript{\textdagger}Corresponding author}
}

\date{}

\begin{document}

\maketitle

%%%%%%%%%%%%%%%%%%%%%%%%%%%%%%%%%%%%%%%%%%%%%%%%%%%%%%%%%%%%%%%%%%%%%%%%
%%%%%%                          ABSTRACT                          %%%%%%
%%%%%%%%%%%%%%%%%%%%%%%%%%%%%%%%%%%%%%%%%%%%%%%%%%%%%%%%%%%%%%%%%%%%%%%%
\begin{abstract}
Autoregressive language models execute Transformer layers sequentially, creating a latency bottleneck that is not removed by conventional tensor or pipeline parallelism. We study whether this layerwise dependency can be relaxed by treating the hidden-state trace across layers as the solution of a nonlinear residual equation and solving it with parallel Newton-style updates. While this view is principled, exact Newton corrections require expensive Jacobian-vector products and naive fixed-point iterations are unstable on trained Transformers.
We introduce Structured Newton Layer Parallelism (SNLP), a training and inference framework that replaces exact layer Jacobians with cheap architecture-induced surrogate dynamics. In residual Transformers, this yields Identity Newton (IDN), where the correction reduces to a prefix-sum-like update; in mHC-style architectures, HC Newton (HCN) uses the model's residual mixing matrix. We also study SNLP-aware training, including pretraining regularization and direct SNLP-forward SFT.
Experiments on Nanochat-scale Transformers show that SNLP exposes a practical speed-quality frontier: on 0.5B models, it reaches up to $2.58\times$ wall-clock speedup, and a less aggressive configuration reaches $1.40\times$ speedup without increasing PPL. The useful tradeoff comes from the biased finite-iteration computation induced by IDN/HCN rather than exact recovery of the sequential trace. We further show that SNLP-forward SFT can preserve downstream task accuracy, and that SNLP can serve as a drafter for self-speculative decoding while a sequential verifier preserves output correctness.\footnote{For a note on changes from the previous arXiv version, see \cref{app:version-note}.} Code is available at \url{https://github.com/phymhan/nanochat-snlp}.
\end{abstract}





%%%%%%%%%%%%%%%%%%%%%%%%%%%%%%%%%%%%%%%%%%%%%%%%%%%%%%%%%%%%%%%%%%%%%%%%
%%%%%%                       Introduction                         %%%%%%
%%%%%%%%%%%%%%%%%%%%%%%%%%%%%%%%%%%%%%%%%%%%%%%%%%%%%%%%%%%%%%%%%%%%%%%%
% To Avert This Unholy Evolutionary Trajectory

\section{Introduction}

Transformer language models~\cite{vaswani2017attention} are sequential in two distinct senses. Token generation is autoregressive~\cite{radford2019language,brown2020language}, but even for a fixed token prefix, the hidden state must normally pass through the network one layer at a time. Tensor parallelism~\cite{shoeybi2019megatron}, pipeline parallelism~\cite{huang2019gpipe}, kernel fusion~\cite{dao2022flashattention}, batching, KV caching~\cite{kwon2023efficient}, and speculative decoding~\cite{leviathan2023fast,chen2023accelerating} improve the efficiency of each layer or token step, but they do not remove the layer dependency chain. As models become deeper~\cite{kaplan2020scaling,touvron2023llama} and decoding remains latency-sensitive, this depthwise dependency becomes a natural target for algorithmic parallelism.

A principled way to expose such parallelism is to view the entire sequence of hidden states across layers as the solution of a nonlinear residual equation. This is analogous to DEER-style~\cite{lim2024parallelizing} parallelization of nonlinear recurrences, where Newton iterations solve for all states in a chain jointly rather than executing the chain strictly left-to-right~\cite{danieli2023deeppcr}. Applied along the depth axis, this perspective suggests that many Transformer layer states could be updated in parallel. However, exact Newton updates require the Jacobian of each full layer with respect to its input. For language-model hidden states, these Jacobians are too large to materialize, and even Jacobian-vector or finite-difference approximations can consume the latency budget that layer parallelism is meant to save. Cheap fixed-point or Jacobi iterations avoid this cost~\cite{song2021accelerating,santilli2023accelerating}, but are often unstable or slow on trained residual networks.

We introduce \textbf{Structured Newton Layer Parallelism} (SNLP), a training and inference framework that makes this Newton view practical by replacing exact layer Jacobians with cheap structured surrogates. In residual Transformers, the identity residual path gives the simplest surrogate, yielding \textbf{Identity Newton} (IDN): the correction reduces to additive prefix-style propagation over depth. For HC/mHC-style models~\cite{zhu2025hyper,xie2025mhc}, the architecture exposes a learned residual mixing matrix, yielding \textbf{HC Newton} (HCN). \textbf{Diagonal Newton} (DiagN) connects SNLP to quasi-DEER and scan-based linear recurrences~\cite{gonzalez2024towards}, but its Jacobian-estimation cost makes practical speedup more demanding. In all cases, the expensive nonlinear layer or chunk forwards are parallelizable, while the Newton correction is a lightweight structured recurrence.

The second ingredient is training co-design. A pretrained sequential model need not be compatible with a cheap surrogate Jacobian, so we study SNLP-aware training in two forms: a pretraining regularizer that matches finite-iteration SNLP states to the sequential trace, and direct SNLP-forward SFT that trains the approximate computation used at inference. These objectives encourage layer dynamics that are easier to solve with the chosen surrogate and allow task adaptation to preserve SNLP-compatible behavior.

Our experiments show that layer-parallel inference can be useful in practice, but not as a universal post-training acceleration trick. On trained-from-scratch Nanochat-scale models, SNLP exposes a speed-quality frontier: aggressive configurations achieve large wall-clock speedups at the cost of PPL loss, while less aggressive configurations recover near-sequential quality. At the 0.5B scale, SNLP reaches up to $2.58\times$ speedup, and one IDN-regularized configuration reaches $1.40\times$ speedup without increasing PPL. At the 3B scale, speedups are harder to realize with our current PyTorch-level implementation, but the IDN-regularized model still shows an encouraging frontier point: $1.20\times$ speedup with a 3.9\% PPL increase.

This behavior is not simply exact numerical recovery of the sequential forward. Exact convergence of the Newton formulation recovers the sequential trace, but practical SNLP uses approximate structured corrections, finite iteration counts, chunking, fusion, and initialization choices. We therefore view IDN and HCN as inducing \textbf{solver-induced inference bias}: the finite-iteration computation differs from strict sequential execution in a way that can expose practical speed-quality tradeoffs and serve as a drafter in verified decoding.
Our contributions are:
\begin{itemize}[nosep,leftmargin=15pt]
    \item We formulate layer-parallel language-model inference as structured surrogate Newton solving over the hidden-state trace, instantiated as IDN, HCN, and DiagN.
    \item We introduce SNLP-aware training, which can enable useful speed-quality tradeoffs with small PPL loss or preserve downstream accuracy under SNLP inference.
    \item We introduce chunkwise layer fusion, which groups multiple depthwise-parallel layers into wider executable chunks before applying the structured Newton correction.
    \item We analyze the resulting solver-induced inference bias through correction ordering, propagation, variance-reduction, and layer-coupling ablations.%, and identify limitations on off-the-shelf pretrained models.
\end{itemize}

\begin{figure*}[t]
  \centering
  \makebox[\textwidth][c]{\includegraphics[width=0.98\textwidth]{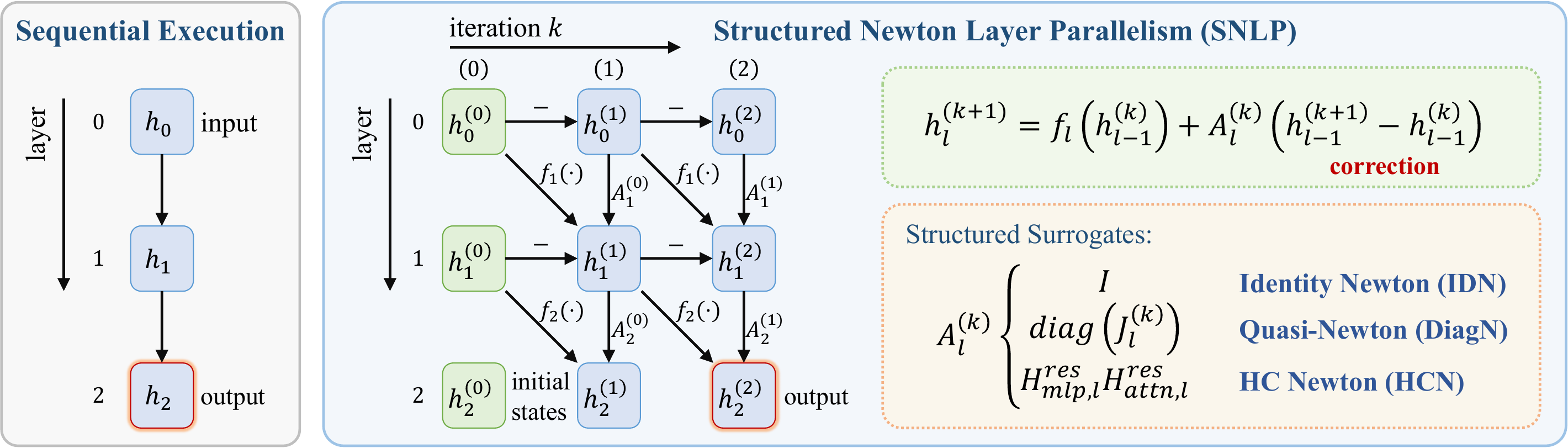}}
  \caption{
  Structured Newton Layer Parallelism (SNLP) replaces sequential layer execution with iterative layer-parallel updates.
  At each iteration, layer states are updated using the block function $f_l$ and a cheap structured Newton surrogate $A_l^{(k)}$, instantiated as IDN, HCN, or DiagN depending on the architecture.
  }
  \label{fig:overview}
  \vspace{-0.2em}
\end{figure*}

\section{Related Work}

\noindent\textbf{Parallel nonlinear solvers.}
SNLP builds on the view that a sequential computation can be solved as a coupled nonlinear system. DEER applies Newton's method to nonlinear recurrences and uses parallel scan to solve the resulting linearized dynamics~\cite{lim2024parallelizing}; later work extends this perspective to MCMC chains~\cite{zoltowski2025parallelizing} and improves stability and scalability with quasi-Newton and Kalman-style approximations~\cite{gonzalez2024towards}. Song et al.~\cite{song2021accelerating} frame feedforward computation as parallel nonlinear equation solving, and Jacobi decoding applies fixed-point iteration to parallelize autoregressive translation~\cite{santilli2023accelerating}. Deep Equilibrium Models~\cite{bai2019deep} take a complementary view, finding fixed points of weight-tied infinite-depth networks via root-finding. Picard iterations parallelize diffusion-model sampling by exploiting identity Jacobian structure near Euler-discretized dynamics~\cite{shih2023parallel}; \citep{gonzalez2026unifying} unify Newton, Picard, and Jacobi under a linear dynamical systems framework and show convergence depends on Jacobian approximation fidelity. Our IDN is equivalent to Picard (identity approximation); DiagN and HCN are progressively richer quasi-Newton variants in this taxonomy. Our work rotates this line of work from sequence length to Transformer depth, and focuses on structured surrogates that avoid full layer Jacobians.

\noindent\textbf{Associative scans and structured recurrences.}
Parallel prefix scan is a classical primitive~\cite{blelloch1990prefix} that has become central to efficient recurrent and state-space models. Linear recurrent networks can be parallelized over sequence length with scan~\cite{martin2018parallelizing}; structured state-space models such as S4~\cite{gu2022efficiently} and Mamba~\cite{gu2024mamba} use related hardware-aware recurrent kernels and scan-style algorithms. SNLP uses the same computational principle for depthwise correction: when the surrogate is identity, diagonal, or a small matrix, the Newton correction becomes a cheap structured recurrence.

\noindent\textbf{Depth mixing and residual architectures.}
Residual connections~\cite{he2016deep} are central to deep Transformer training, and several architectures modify how information flows across depth. Hyper-Connections and mHC introduce learned residual-stream mixing and stabilization mechanisms~\cite{zhu2025hyper,xie2025mhc}; AttnRes replaces fixed residual accumulation with learned attention over previous layer outputs~\cite{chen2026attention}. Value residual learning and x0-style residual connections also alter how features persist through depth~\cite{zhou2025value,modded_nanogpt_2024}. Weight-tied and looped architectures, including Universal Transformers~\cite{dehghani2019universal}, ALBERT~\cite{lan2020albert}, recurrent-depth models~\cite{geiping2025scaling}, and Hyperloop Transformers~\cite{zeitoun2026hyperloop}, reuse layers across depth. SNLP is complementary: rather than only changing the forward architecture, it asks whether the resulting depth dynamics expose a cheap surrogate for Newton-style layer-parallel inference.

\noindent\textbf{Efficient language-model inference.}
Most efficient LLM inference work accelerates token-level decoding through batching, KV caching~\cite{kwon2023efficient}, quantization, memory-aware execution~\cite{alizadeh2024llm}, kernel engineering~\cite{dao2022flashattention}, speculative decoding~\cite{leviathan2023fast,chen2023accelerating}, early exit~\cite{schuster2022confident}, or serving systems~\cite{zhen2025taming,miao2025towards}. These techniques improve the execution of the standard sequential layer stack, whereas SNLP targets a different bottleneck: the dependency chain across layers for a fixed token prefix. Our experiments use Nanochat as a compact from-scratch training and evaluation harness~\cite{nanochat}; we also run preliminary post-hoc and finetuning experiments on representative open-weight decoder-only models, including Qwen2.5 and TinyLlama~\cite{qwen2.5,zhang2024tinyllama}. The gap between trained-from-scratch and off-the-shelf results suggests that layer-parallel inference benefits from training/inference co-design, leaving stronger pretrained-model adaptation to future work.

\noindent\textbf{Layer-parallel computation.}
\citep{jiang2026layer} apply multigrid-in-time to parallelize Transformer \emph{training} across layers via a neural ODE formulation, complementing our inference-side approach. \citep{calvo2026leveraging} parallelize consecutive layer pairs post hoc by dropping inter-layer dependencies; SNLP instead applies Newton corrections and co-designs training with inference.

\section{Methods}
\label{sec:method}

\subsection{Background}

\noindent\textbf{Layer traces as residual equations.}
Consider a depth-$L$ model with hidden states $h_0,\ldots,h_L$ and layer maps
\begin{equation}
    h_l = f_l(h_{l-1}), \qquad l=1,\ldots,L .
\end{equation}
Here $l$ indexes depth, while superscripts such as $(k)$ will index iterative solver steps. Rather than viewing the forward pass only as a sequential program, we can view the entire hidden-state trace $\mathbf h=(h_1,\ldots,h_L)$ as the solution of a nonlinear residual equation. Define
\begin{equation}
    \label{eq:residual}
    G_l(\mathbf h) = h_l - f_l(h_{l-1}), \qquad
    G(\mathbf h) = (G_1(\mathbf h),\ldots,G_L(\mathbf h)).
\end{equation}
The usual sequential forward pass is exactly the zero-residual trace $G(\mathbf h)=0$. This formulation exposes a different source of parallelism: instead of computing layers one after another, one may iteratively solve for all layer states jointly.

\noindent\textbf{Newton-style updates over depth.}
DEER applies Newton's method to nonlinear recurrences by linearizing the transition at the current iterate and solving the resulting linear recurrence in parallel~\cite{lim2024parallelizing,zoltowski2025parallelizing}. Rotating this view by $90^\circ$, the depth axis of any block-sequential model--a Transformer, CNN, or recurrent stack--can be treated as the recurrence axis. At solver iteration $k$, the exact Newton update over layers can be written as
\begin{equation}
    h_l^{(k+1)}
    =
    f_l\!\left(h_{l-1}^{(k)}\right)
    +
    J_l^{(k)}
    \left(h_{l-1}^{(k+1)} - h_{l-1}^{(k)}\right),
    \qquad
    J_l^{(k)} =
    \frac{\partial f_l}{\partial h_{l-1}}\!\left(h_{l-1}^{(k)}\right).
\end{equation}
This recurrence is equivalent to applying Newton's method to the stacked residual system in \Eqref{eq:residual} because the residual Jacobian is block lower-bidiagonal; we refer readers to prior derivations of this equivalence in DEER-style solvers~\cite{zoltowski2025parallelizing,gonzalez2024towards}. The challenge is that $J_l^{(k)}$ is the Jacobian of an entire layer or block output with respect to its input. For language-model hidden states, materializing this operator is infeasible, and even Jacobian-vector products or finite-difference approximations can consume the latency budget that layer parallelism is meant to save. Naive fixed-point updates avoid this cost but are often unstable on trained residual networks. The practical question is therefore whether we can replace the exact layer Jacobian with a cheap structured surrogate that preserves enough of the Newton correction to make finite-iteration, layer-parallel inference useful.

\subsection{Structured Newton Layer Parallelism}

SNLP replaces the exact layer Jacobian in \Eqref{eq:residual} with a cheap structured surrogate. Let the first $S$ layers be evaluated sequentially, producing a prefix state $h_S$. The remaining suffix $\{S+1,\ldots,L\}$ is solved by iterative correction. At iteration $k$, each suffix layer is first evaluated using the current estimate of its input,
\begin{equation}
    \tilde h_l^{(k)} = f_l\!\left(h_{l-1}^{(k)}\right),
    \qquad l=S+1,\ldots,L .
\end{equation}
These evaluations are independent across $l$ and can be batched or fused. SNLP then applies the structured Newton correction
\begin{equation}
    \label{eq:snlp-update}
    h_l^{(k+1)}
    =
    \tilde h_l^{(k)}
    +
    A_l^{(k)}
    \left(h_{l-1}^{(k+1)} - h_{l-1}^{(k)}\right),
    \qquad h_S^{(k+1)}=h_S ,
\end{equation}
where $A_l^{(k)}$ is a surrogate for the exact block Jacobian $J_l^{(k)}$. If $A_l^{(k)}=J_l^{(k)}$, this recovers the exact DEER/Newton update over depth. SNLP instead chooses $A_l^{(k)}$ so that the correction is much cheaper than evaluating or materializing the true Jacobian, while still propagating information from earlier corrected layer states to later ones.

The update in \Eqref{eq:snlp-update} separates the two costs that matter for inference. The nonlinear layer evaluations $\tilde h_l^{(k)}$ are parallel across the suffix and dominate GPU work. The Newton correction still propagates through depth, but because $A_l^{(k)}$ is either trivial to compute or directly available from the architecture, this sequential correction is cheap relative to a Transformer block. Thus SNLP realizes speedup by parallelizing the expensive block forwards while keeping only a lightweight structured recurrence on the critical path. After $K$ iterations, the model projects the final corrected state $h_L^{(K)}$ to logits.

\noindent\textbf{Effect of the correction.}
The correction in \Eqref{eq:snlp-update} is what moves information across the whole suffix within a single solver iteration. Once the layer outputs $\tilde h_l^{(k)}$ are computed, the corrected prefix state propagates from layer $S$ to layer $L$ through the structured recurrence, so $h_L^{(k+1)}$ depends on the corrections from all layers $S+1,\ldots,L$. Without this correction, a naive parallel fixed-point update only advances information by one layer per iteration: after $K$ iterations, the effect of the prefix can reach only the next $K$ layers of the suffix.
We verify this propagation effect empirically in \cref{sec:inference_effect} and \cref{app:correction-propagation}.

\subsection{Structured Surrogates}

\noindent\textbf{Identity Newton (IDN).}
For residual Transformer blocks, $f_l(x)$ contains an explicit identity path. SNLP uses the architecture-induced surrogate
\begin{equation}
    A_l^{(k)} = I .
\end{equation}
The correction becomes
\begin{equation}
    h_l^{(k+1)}
    =
    \tilde h_l^{(k)}
    +
    h_{l-1}^{(k+1)} - h_{l-1}^{(k)} ,
\end{equation}
which reduces the Newton correction to additive propagation of the previous-layer correction. This is our main residual-Transformer instantiation because it requires no Jacobian estimation and makes the correction essentially a prefix-sum over depth. We refer to this variant as Identity Newton (IDN).

\noindent\textbf{HC Newton (HCN).}
For hyper-connection and mHC-style models~\cite{zhu2025hyper,xie2025mhc}, the architecture exposes an explicit residual mixing matrix over streams. If a block applies residual mixing matrices $H^{\mathrm{res}}_{\mathrm{attn},l}$ and $H^{\mathrm{res}}_{\mathrm{mlp},l}$, we use
\begin{equation}
    A_l = H^{\mathrm{res}}_{\mathrm{mlp},l} H^{\mathrm{res}}_{\mathrm{attn},l}.
    \label{eq:hcn-A}
\end{equation}
This surrogate is small: it acts on the stream dimension rather than on the full hidden dimension. The mHC case demonstrates that SNLP is not tied to the identity residual path; any architecture with a cheap structured approximation to inter-layer sensitivity can define an SNLP correction.

\noindent\textbf{Diagonal Newton (DiagN).}
A closer approximation to the exact Newton step uses only the diagonal of the layer Jacobian,
\begin{equation}
    A_l^{(k)} = \operatorname{diag}\!\left(J_l^{(k)}\right).
\end{equation}
This connects SNLP to quasi-DEER and ELK-style approximations~\cite{gonzalez2024towards}. With a diagonal surrogate, the correction in \Eqref{eq:snlp-update} becomes an elementwise affine recurrence over depth and can be evaluated efficiently by an associative prefix scan~\cite{blelloch1990prefix,martin2018parallelizing,gu2024mamba}. In our implementation, the diagonal can be estimated by a Hutchinson-style finite-difference or VJP estimator~\cite{hutchinson1990stochastic,zoltowski2025parallelizing,bekas2007estimator}, optionally only on a subset of layers.

\subsection{SNLP-Aware Training}

Off-the-shelf sequential models need not have layer dynamics that match a cheap surrogate. We use SNLP-aware training in two ways: an auxiliary matching loss during pretraining, and direct SNLP-forward SFT during task adaptation. For pretraining, we add a loss that makes a finite SNLP solve match the sequential trace. For each suffix length $N\in\mathcal S$, let $\mathcal T_N$ be the stride-selected supervised layers in that suffix, and let $\hat h_l^{\mathrm{SNLP}}(N,K;A)$ be the SNLP state at layer $l$ after $K$ iterations with surrogate family $A$. We optimize
\begin{equation}
    \mathcal L
    =
    \mathcal L_{\mathrm{CE}}
    +
    \lambda
    \sum_{N\in\mathcal S}
    \sum_{l\in\mathcal T_N}
    \frac{
        \left\|\hat h_l^{\mathrm{SNLP}}(N,K;A)-h_l^{\mathrm{seq}}\right\|_2
    }{
        \left\|h_l^{\mathrm{seq}}\right\|_2+\epsilon
    } .
    \label{eq:snlp-training-loss}
\end{equation}
In our runs, $K=1$ during training and $\mathcal S$ contains one or more configured suffix lengths. The set $\mathcal T_N$ controls where the matching loss is applied: stride 0 uses only the final layer, $\mathcal T_N=\{L\}$, while positive strides add sparse intermediate layers and always include $L$ to reduce memory cost; see \cref{tab:idn_reg_ablation} for ablations. The surrogate $A$ is identity for IDN, the stream-mixing matrix for HCN, and diagonal for DiagN. This objective does not make layers removable; rather, it makes the chosen structured correction a better finite-iteration solver for the sequential trace. For SNLP-forward SFT, we instead use the SNLP computation itself as the model forward pass and optimize the standard supervised loss, allowing task adaptation to the approximate layer-parallel computation used at inference.

\subsection{Inference With Fusion and Chunking}

At inference time, SNLP runs a sequential prefix and applies \Eqref{eq:snlp-update} to a suffix of $N=L-S$ layers. The suffix hidden states can be initialized from the prefix state $h_S$, from a one-shot batched forward, or from a lightweight predictor; our main evaluations focus on simple prefix-state and batched-forward initializations. The number of iterations $K$ controls the quality-cost tradeoff.

\noindent\textbf{Layer fusion.}
Wall-clock speedups require more than replacing the Jacobian. We therefore combine SNLP correction with GPU-oriented execution of the suffix. In the batched form, per-layer weights are stacked so all suffix layers evaluate in one grouped operation. In the fused form, several layers that read the same input are combined into one wider layer: the attention $Q,K,V$ projections and MLP expansion matrices are concatenated along their output dimension, while the attention output projection and MLP down-projection are concatenated along their input dimension. Equivalently, the fused layer computes all branch outputs in one wide matmul and performs the required sum-reductions after the attention output projection and after the MLP projection. This converts layer-parallel algorithmic structure into larger GPU-efficient matrix multiplies.

\noindent\textbf{Chunkwise strategy.}
For more aggressive parallelization, we split the suffix into multiple fused chunks, inspired by DeltaNet-style chunkwise parallelization~\cite{yang2024deltanet}. Each chunk is treated as a wide layer as above, and all chunk forwards are parallelizable because they use the current chunk-input estimates from iteration $k$. SNLP then applies the structured Newton correction between chunk outputs rather than between individual layers:
\begin{equation}
    h_c^{(k+1)}
    =
    \tilde h_c^{(k)}
    +
    A_c^{(k)}
    \left(h_{c-1}^{(k+1)} - h_{c-1}^{(k)}\right),
    \label{eq:snlp-iterative}
\end{equation}
where $c$ indexes chunks and $A_c^{(k)}$ is the corresponding identity or architecture-induced chunk surrogate. Chunking trades a coarser solver approximation for better hardware utilization, since the expensive work is executed as a small number of wide parallel chunk forwards followed by a cheap correction across chunks. These fusion choices change the finite-iteration computation, so the resulting model should be understood as practical SNLP inference rather than exact recovery of the sequential forward.

\section{Analysis}
\label{sec:analysis}
Exact convergence of Newton's method on \Eqref{eq:residual} recovers the sequential forward pass, but practical SNLP uses approximate surrogates, finite iterations, initialization, fusion, and chunking; together these define a solver-induced inference bias. We summarize the main mechanisms here and defer derivations to the appendix.

\noindent\textbf{Training-side effects.}
SNLP-aware training makes a cheap structured correction match the sequential final state. For residual blocks $f_l(x)=x+g_l(x)$, IDN training encourages $g_l(h_S)\approx g_l(h_l^{\mathrm{seq}})$ over the suffix, putting implicit Lipschitz pressure on the non-residual branch: smaller $J_{g_l}$ makes $J_{f_l}=I+J_{g_l}$ closer to the IDN surrogate. This encourages suffix layers to act more like stable feature-correction modules. The suffix is still useful, but its features become less sensitive to the exact state where they are evaluated.

\noindent\textbf{Inference-side effects.}
IDN and fused SNLP evaluate many suffix contributions at a common or chunk-level input instead of along the fully accumulated sequential chain. This removes part of the variance from layerwise error compounding, at the cost of bias from evaluating $g_l(h_S)$ rather than $g_l(h_l^{\mathrm{seq}})$. When SNLP training keeps this bias small, the variance reduction helps explain why finite-iteration SNLP can stay near the sequential model while changing the computation enough to expose latency tradeoffs. Fusion and chunking add another bias: fused branches see summed cross-layer signals, which can alter feature interactions and can hurt when chunks are too aggressive.

\noindent\textbf{Layer coupling.}
The results suggest that SNLP benefits from structured depth coupling rather than removing depth interactions entirely. HC and mHC parameterize residual-stream mixing~\cite{zhu2025hyper,xie2025mhc}, while AttnRes learns attention over depth~\cite{chen2026attention}. SNLP is complementary: IDN uses identity coupling, HCN uses the learned residual mixing matrix, and fused SNLP induces implicit cross-layer coupling inside each chunk.

\newcommand{\modelppl}[2]{%
\begin{tabular}[t]{@{}l@{}}
#1\\[-0.2ex]
{\small [#2]}
\end{tabular}%
}

\begin{table*}[t]
\centering
\resizebox{\linewidth}{!}{%
\begin{tabular}{lllrrrrrrr}
\toprule
Model & Config & $K$ & PPL & $\Delta$PPL & Speedup & Top-1 & LogitSim & EmbSim & AR Match \\
\midrule

\multicolumn{10}{c}{\textbf{Nanochat-3B standard}} \\
\midrule
\multirow{2}{*}{\modelppl{No Reg.}{10.10}}
& {\small\texttt{8xF1-h0}} & 4 & 10.65 & \bad{+5.5\%} & \comp{$0.95\times$} & 85.8\% & 0.993 & 0.965 & 61.7\% \\
& {\small\texttt{8xF1-h0}} & 8 & 10.13 & \comp{+0.3\%} & \bad{$0.77\times$} & 99.4\% & 1.000 & 0.996 & 95.9\% \\
\cmidrule(lr){1-10}

\multirow{3}{*}{\modelppl{IDN Reg.}{10.07 (\good{-0.3\%})}}
& {\small\texttt{8xF1-h0}} & 1 & 10.46 & \bad{+3.9\%} & \good{$1.20\times$} & 84.0\% & 0.985 & 0.961 & 55.0\% \\
& {\small\texttt{8xF1-fwd}} & 1 & 10.43 & \bad{+3.6\%} & \good{$1.10\times$} & 85.0\% & 0.988 & 0.963 & 58.0\% \\
& {\small\texttt{8xF1-h0}} & 4 & 10.09 & \comp{+0.2\%} & \comp{$0.95\times$} & 99.1\% & 1.000 & 0.995 & 95.2\% \\

\midrule
\multicolumn{10}{c}{\textbf{Nanochat-0.5B standard}} \\
\midrule
\multirow{2}{*}{\modelppl{No Reg.}{15.21}}
& {\small\texttt{8xF1-h0}} & 4 & 15.42 & \comp{+1.4\%} & \comp{$0.99\times$} & 95.2\% & 1.000 & 0.993 & 90.1\% \\
& {\small\texttt{8xF1-fwd}} & 4 & 15.35 & \comp{+0.9\%} & \bad{$0.93\times$} & 97.8\% & 1.000 & 0.997 & 95.1\% \\
\cmidrule(lr){1-10}

\multirow{3}{*}{\modelppl{IDN Reg.}{15.36 (\comp{+1.0\%})}}
& {\small\texttt{24xF1-h0}} & 2 & 18.09 & \bad{+17.8\%} & \good{$1.88\times$} & 74.3\% & 0.989 & 0.969 & 54.5\% \\
& {\small\texttt{24xF1-h0}} & 4 & 16.42 & \bad{+6.9\%} & \good{$1.33\times$} & 90.0\% & 0.998 & 0.987 & 81.4\% \\
& {\small\texttt{12xF1-fwd}} & 2 & 15.59 & \comp{+1.5\%} & \good{$1.14\times$} & 95.0\% & 0.999 & 0.993 & 90.8\% \\

\midrule
\multicolumn{10}{c}{\textbf{Nanochat-0.5B w/o x0ve}} \\
\midrule
\multirow{2}{*}{\modelppl{No Reg.}{17.65}}
& {\small\texttt{20xF1-fwd}} & 4 & 18.91 & \bad{+7.2\%} & \good{$1.27\times$} & 90.3\% & 0.999 & 0.989 & 86.1\% \\
& {\small\texttt{8xF1-h0}} & 4 & 17.75 & \comp{+0.6\%} & \comp{$0.98\times$} & 99.0\% & 1.000 & 0.999 & 98.4\% \\
\cmidrule(lr){1-10}

\multirow{3}{*}{\modelppl{IDN Reg.}{17.57 (\good{-0.5\%})}}
& {\small\texttt{12xF2-h0}} & 1 & 20.55 & \bad{+17.0\%} & \good{\textbf{$2.58\times$}} & 70.5\% & 0.996 & 0.955 & 56.4\% \\
& {\small\texttt{12xF2-h0}} & 2 & 18.46 & \bad{+5.1\%} & \good{$2.09\times$} & 79.4\% & 0.998 & 0.966 & 80.2\% \\
& {\small\texttt{1xF12-h0}} & 1 & 17.56 & \good{-0.1\%} & \good{$1.40\times$} & 87.7\% & 0.999 & 0.976 & 100.0\% \\

\midrule
\multicolumn{10}{c}{\textbf{Nanochat-0.5B-mHC}} \\
\midrule
\multirow{2}{*}{\modelppl{No Reg.}{15.16}}
& {\small\texttt{8xF1-fwd}} & 2 & 15.65 & \bad{+3.2\%} & \good{$1.10\times$} & 93.4\% & 0.998 & 0.991 & 90.9\% \\
& {\small\texttt{8xF1-h0}} & 4 & 15.38 & \comp{+1.5\%} & 1.00$\times$ & 98.2\% & 1.000 & 0.997 & 95.9\% \\
\cmidrule(lr){1-10}

\multirow{3}{*}{\modelppl{HCN Reg.}{15.52 (\comp{+2.4\%})}}
& {\small\texttt{16xF1-h0}} & 1 & 16.93 & \bad{+9.1\%} & \good{$1.60\times$} & 81.4\% & 0.993 & 0.980 & 72.0\% \\
& {\small\texttt{16xF1-h0}} & 4 & 15.71 & \comp{+1.2\%} & \good{$1.14\times$} & 97.3\% & 1.000 & 0.995 & 94.4\% \\
& {\small\texttt{8xF1-h0}} & 4 & 15.60 & \comp{+0.5\%} & 1.00$\times$ & 98.3\% & 1.000 & 0.997 & 95.9\% \\

\bottomrule
\end{tabular}%
}
\caption{Main results for layer-parallel decoding across Nanochat model variants using 2048-token PPL evaluation. Brackets under model names give sequential PPL and, when shown, the change from the corresponding No Reg. baseline. Entries marked gray denote comparable quality or speed: PPL increase within 2.5\%, or speedup at least $0.95\times$ but below sequential. Speed-oriented configurations reach up to $2.58\times$ practical speedup, while quality-oriented configurations recover near-sequential PPL with high LogitSim, EmbSim, and AR Match. AR Match uses 32 generated tokens, so 50\% observed match requires roughly 95\% conditional per-token agreement (\cref{app:ar-match-rate}).}
\label{tab:main_results}
\vspace{-0.2em}
\end{table*}

\begin{table*}[h]
\centering
\scriptsize
\resizebox{\linewidth}{!}{%
\begin{tabular}{lllrrrr}
\toprule
Model & SFT Method & Inference Method & ARC-Easy & ARC-Challenge & MMLU & Avg \\
\midrule

\multicolumn{7}{c}{\textbf{Nanochat-3B standard}} \\
\midrule
\multirow{4}{*}{\shortstack[l]{No Reg.\\[-0.4ex]{\scriptsize [10.10]}}}
& {\tiny\texttt{Sequential}} & {\tiny\texttt{Sequential}} & 74.91\% & 58.86\% & 39.78\% & \textbf{57.85\%} \\
& {\tiny\texttt{Sequential}} & {\tiny\texttt{IDN-8xF1-K1}} & 70.70\% & 54.18\% & 38.93\% & 54.60\% \\
& {\tiny\texttt{IDN-8xF1-K1}} & {\tiny\texttt{Sequential}} & 72.46\% & 51.84\% & 39.71\% & 54.67\% \\
& {\tiny\texttt{IDN-8xF1-K1}} & {\tiny\texttt{IDN-8xF1-K1}} & 73.33\% & 54.85\% & 39.71\% & 55.96\% \\
\cmidrule(lr){1-7}

\multirow{4}{*}{\shortstack[l]{IDN Reg.\\[-0.4ex]{\scriptsize [10.07]}}}
& {\tiny\texttt{Sequential}} & {\tiny\texttt{Sequential}} & 74.74\% & 57.19\% & 40.30\% & \textbf{57.41\%} \\
& {\tiny\texttt{Sequential}} & {\tiny\texttt{IDN-8xF1-K1}} & 74.39\% & 56.86\% & 39.84\% & 57.03\% \\
& {\tiny\texttt{IDN-12xF1-K1}} & {\tiny\texttt{Sequential}} & 70.88\% & 55.52\% & 37.95\% & 54.78\% \\
& {\tiny\texttt{IDN-12xF1-K1}} & {\tiny\texttt{IDN-12xF1-K1}} & 70.88\% & 55.85\% & 38.08\% & 54.94\% \\

\midrule
\multicolumn{7}{c}{\textbf{Nanochat-0.5B standard}} \\
\midrule
\multirow{4}{*}{\shortstack[l]{No Reg.\\[-0.4ex]{\scriptsize [15.21]}}}
& {\tiny\texttt{Sequential}} & {\tiny\texttt{Sequential}} & 44.91\% & 39.13\% & 31.03\% & \textbf{38.36\%} \\
& {\tiny\texttt{Sequential}} & {\tiny\texttt{IDN-12xF1-K2}} & 31.40\% & 30.43\% & 28.22\% & 30.02\% \\
& {\tiny\texttt{IDN-12xF1-K2}} & {\tiny\texttt{Sequential}} & 38.42\% & 31.44\% & 31.74\% & 33.87\% \\
& {\tiny\texttt{IDN-12xF1-K2}} & {\tiny\texttt{IDN-12xF1-K2}} & 39.30\% & 32.11\% & 31.74\% & 34.38\% \\
\cmidrule(lr){1-7}

\multirow{4}{*}{\shortstack[l]{IDN Reg.\\[-0.4ex]{\scriptsize [15.36]}}}
& {\tiny\texttt{Sequential}} & {\tiny\texttt{Sequential}} & 37.72\% & 38.80\% & 31.94\% & 36.15\% \\
& {\tiny\texttt{Sequential}} & {\tiny\texttt{IDN-12xF1-K2}} & 38.25\% & 37.79\% & 31.29\% & 35.78\% \\
& {\tiny\texttt{IDN-12xF1-K2}} & {\tiny\texttt{Sequential}} & 43.16\% & 35.79\% & 30.50\% & \textbf{36.48\%} \\
& {\tiny\texttt{IDN-12xF1-K2}} & {\tiny\texttt{IDN-12xF1-K2}} & 42.28\% & 35.79\% & 30.63\% & 36.23\% \\
\bottomrule
\end{tabular}%
}
% \vspace{-0em}
\caption{Downstream multiple-choice accuracy after SFT with either sequential or SNLP forward. Each checkpoint is evaluated with both sequential and IDN inference.}
\vspace{-1em}
\label{tab:sft_task_results}
\end{table*}

\section{Experiments}
\label{sec:experiments}

\noindent\textbf{From-scratch models.}
Our main results use Nanochat~\cite{nanochat} models trained from scratch at two 32-layer scales: a 3B model with $n_{\mathrm{embd}}=2048$ and 16 heads, and a 0.5B model with $n_{\mathrm{embd}}=640$ and 5 heads. Forward Nanochat uses rotary position embeddings~\cite{su2021roformer}, an $x_0$ residual connection~\cite{modded_nanogpt_2024}, and value embeddings~\cite{zhou2025value}. For both standard scales, we train No Reg., IDN Reg., and DiagN Reg. variants. We also train 0.5B variants without $x_0$/VE to isolate SNLP from Nanochat-specific features, and an mHC model~\cite{zhu2025hyper,xie2025mhc}, where HCN uses the learned matrix surrogate.

\noindent\textbf{Off-the-shelf models.}
We also evaluate SNLP post hoc on Qwen2.5-0.5B-Instruct~\cite{qwen2.5} and TinyLlama-1.1B-Chat-v1.0~\cite{zhang2024tinyllama}. Their best SNLP configurations can match sequential perplexity and are slower than sequential execution; see \cref{tab:ots_models}. We additionally run a preliminary TinyLlama finetuning study to test whether SNLP compatibility can be introduced after pretraining.

\noindent\textbf{Inference configurations.}
We evaluate chunkwise SNLP with IDN correction for residual models and HCN correction for mHC models. A configuration \texttt{NxFM-init} denotes $N$ parallel chunks, each fusing $M$ layers, with initialization \texttt{h0} from the prefix state $h_S$ or \texttt{fwd} from a one-shot parallel forward $f_l(h_S)$. For example, \texttt{8xF2-fwd} uses 8 fused 2-layer chunks with one-shot initialization. Preheat initialization is deferred to the appendix.

\noindent\textbf{Metrics and protocol.}
All models are implemented in PyTorch~\cite{paszke2019pytorch} using HuggingFace Transformers~\cite{wolf2020transformers}. Training and evaluation use ClimbMix dataset~\cite{diao2025climb}; PPL is measured on a fixed 1M-token validation split. Timing uses batch size 1 on H100 GPUs with 50 warmup and 200 measured runs. Top-1 and LogitSim are prefill metrics against the original sequential model. AR Match is token-level greedy-generation agreement over 32 tokens; for fused-weight variants, it compares to sequential execution of the same fused model. EmbSim compares generated text to original sequential generation using BGE-small-en-v1.5 embeddings~\cite{bge_embedding}. For downstream evaluation, we follow Nanochat~\cite{nanochat} to SFT on SmolTalk~\cite{allal2025smollm2} and evaluate multiple-choice accuracy on ARC~\cite{clark2018arc} and MMLU~\cite{hendrycks2021mmlu}.
% \vspace{-1em}

\subsection{Main results}
\noindent\textbf{Pretraining evaluation.}
\cref{tab:main_results} reports two SNLP configurations for each base model and three for each SNLP-regularized model; checkpoint settings are listed in \cref{tab:ckpt_selection}. Within each model variant,
\begin{wraptable}{r}{0.58\linewidth}
% \vspace{-2.5em}
\centering
\scriptsize
\resizebox{\linewidth}{!}{%
\begin{tabular}{lrrrrr}
\toprule
Config & $s$ & $B$ & $\alpha$ & Tok/Rnd & Speedup \\
\midrule
\multicolumn{6}{c}{\textbf{Nanochat-0.5B standard, IDN Reg.}} \\
\midrule
\multirow{3}{*}{{\tiny\texttt{24xF1-K2-h0}}} & \multirow{3}{*}{$1.86\times$} & 2 & 0.969 & 2.94 & $1.401\times$ \\
& & \textbf{4} & 0.963 & 4.87 & \textbf{$1.475\times$} \\
& & 8 & 0.952 & 8.66 & $1.408\times$ \\
\cmidrule(lr){1-6}
\multirow{3}{*}{{\tiny\texttt{24xF1-K4-h0}}} & \multirow{3}{*}{$1.30\times$} & 2 & 0.999 & 3.00 & $1.181\times$ \\
& & 4 & 0.997 & 5.01 & $1.221\times$ \\
& & \textbf{8} & 0.995 & 9.02 & \textbf{$1.236\times$} \\
\midrule
\multicolumn{6}{c}{\textbf{Nanochat-0.5B w/o x0ve, IDN Reg.}} \\
\midrule
\multirow{3}{*}{{\tiny\texttt{12xF2-K1-h0}}} & \multirow{3}{*}{$2.63\times$} & 2 & 0.925 & 2.85 & $1.578\times$ \\
& & \textbf{4} & 0.914 & 4.67 & \textbf{$1.670\times$} \\
& & 8 & 0.889 & 8.16 & $1.456\times$ \\
\cmidrule(lr){1-6}
\multirow{3}{*}{{\tiny\texttt{12xF2-K2-h0}}} & \multirow{3}{*}{$2.15\times$} & 2 & 0.965 & 2.93 & $1.498\times$ \\
& & \textbf{4} & 0.958 & 4.84 & \textbf{$1.604\times$} \\
& & 8 & 0.942 & 8.58 & $1.514\times$ \\
\midrule
\multicolumn{6}{c}{\textbf{Nanochat-0.5B-mHC, HCN Reg.}} \\
\midrule
\multirow{3}{*}{{\tiny\texttt{16xF1-K1-h0}}} & \multirow{3}{*}{$1.63\times$} & 2 & 0.998 & 3.00 & $1.343\times$ \\
& & 4 & 0.997 & 5.00 & $1.436\times$ \\
& & \textbf{8} & 0.995 & 9.01 & \textbf{$1.481\times$} \\
\cmidrule(lr){1-6}
\multirow{3}{*}{{\tiny\texttt{16xF1-K2-h0}}} & \multirow{3}{*}{$1.44\times$} & 2 & 0.999 & 3.00 & $1.252\times$ \\
& & 4 & 0.999 & 5.01 & $1.316\times$ \\
& & \textbf{8} & 0.997 & 9.04 & \textbf{$1.341\times$} \\
\bottomrule
\end{tabular}%
}
\caption{Self-speculative decoding with SNLP drafters and sequential verification. \(s\) is drafter forward speedup, \(B\) is draft block size, \(\alpha\) is greedy acceptance rate, Tok/Rnd is tokens per SSD round, and Speedup is theoretical end-to-end speedup.}
\label{tab:ssd_results}
\vspace{-2em}
\end{wraptable}
rows are ordered from higher PPL to lower PPL, exposing the expected speed-quality tradeoff: more aggressive layer parallelism gives larger speedup, while additional Newton iterations recover quality at lower speed. On the small 0.5B w/o $x_0$/VE IDN-regularized model, SNLP reaches $1.40\times$ speedup without increasing PPL, or up to $2.58\times$ speedup with 17.0\% PPL loss. This makes SNLP suitable for settings that can tolerate quality degradation in exchange for latency, such as speculative decoding, which we study later. The 3B models are harder to accelerate in our implementation. At this width, sequential Transformer blocks already saturate the H100 more effectively, so PyTorch-level layer fusion does not overcome the overheads. Custom fused kernels or software-hardware co-design, such as compute-in-memory-style execution~\cite{wan2022compute}, may be needed to realize the algorithmic parallelism at larger scale.

The diagonal variant is useful as an ablation because it relaxes the strong identity assumption, but it adds extra block evaluations or autodiff work. \Cref{tab:diag_jacobian_results} shows that, unlike IDN inference, most quality-oriented diagonal-correction configurations only recover comparable PPL to sequential inference. Compared with IDN, DiagN introduces a different solver-induced bias that is numerically closer to the sequential computation, but this does not translate into a better practical speed-quality frontier in our implementation.

\noindent\textbf{Downstream task evaluation.}
\Cref{tab:sft_task_results} evaluates downstream accuracy for different SFT and inference combinations. On 3B, sequential SFT from the IDN-regularized model followed by IDN inference achieves 57.03\% average accuracy, close to 57.85\% from sequential SFT and sequential inference on the base model, and better than 54.60\% from base-model sequential SFT with IDN inference. SNLP-aware SFT on this checkpoint works with \texttt{IDN-12xF1-K1}; using the same \(N=8\) parallelism as pretraining starts with too small a loss, so the SFT signal is too weak to learn a discriminative task head. On 0.5B, IDN-forward SFT with IDN inference reaches 36.23\%, below the 38.36\% base sequential baseline but far above 30.02\% from base-model sequential SFT with IDN inference; this uses more aggressive \texttt{IDN-12xF1-K2} parallelism. Overall, IDN-regularized pretraining better preserves parallel behavior after SFT: switching a sequentially SFT'd IDN-regularized model to IDN inference causes little accuracy drop compared with the corresponding base-model switch.

\noindent\textbf{Self-speculative decoding.}
We also use SNLP as a drafter inside self-speculative decoding, with the same model's sequential forward as the verifier~\cite{leviathan2023fast,chen2023accelerating}. The verifier guarantees identical output to sequential greedy decoding, while SNLP only changes the proposal cost. For block size \(B\), drafter forward speedup \(s\), and per-draft greedy acceptance rate \(\alpha\), we estimate ideal speedup as \(((1-\alpha^{B+1})/(1-\alpha))/(B/s+1)\). \Cref{tab:ssd_results} evaluates 128 validation prompts with prefill length 16 and 1024 greedy generated tokens. We report greedy results because sampling substantially reduces acceptance rates in our runs. The best configuration reaches \(1.67\times\) estimated speedup: a fast \texttt{12xF2} drafter compensates for lower acceptance. The optimal block size is usually \(B=4\), while very high-acceptance mHC and higher-\(K\) IDN configs prefer \(B=8\). Increasing Newton iterations improves \(\alpha\) but slows the drafter, so the net speedup can decrease. HCN maintains especially high greedy acceptance despite nonzero PPL gap, suggesting that it preserves token ranking well.

\subsection{Training effect}

Moderate IDN/HCN regularization usually adds a small amount of pretraining PPL loss, with sequential PPL changes ranging from -0.5\% to +2.4\% in \cref{tab:main_results}. However, as shown in the SFT results above, IDN-regularized models can maintain their parallel behavior after standard SFT.
The diagnostics and inference-effect ablations in this section use a more parallelizable 0.5B IDN checkpoint from \cref{tab:idn_reg_ablation} (\(N=24\), \(\lambda=0.5\), stride 3, no detach; PPL 16.81), rather than the main selected 0.5B IDN checkpoint.
\cref{tab:jacobian_norms} measures the Jacobian of the non-residual branch $g_l$ for the last 8 layers of the 0.5B standard models. For layers 24--30, IDN Reg. reduces spectral estimates by roughly $12\times$ and Hutchinson Frobenius estimates by roughly $12\times$, supporting the implicit Lipschitz-regularization interpretation. \cref{tab:amplification} shows the same reduction in full-layer amplification $\|J_{f_l}v\|/\|v\|$, including removal of the layer-31 outlier.
For diagonal-Jacobian inference, VJP-based estimates may introduce additional bias, especially in HC/mHC-style models where asymmetric routing makes $J^\top v$ deviate from the required forward sensitivity.

\begin{table*}[t]
\centering
\footnotesize
\resizebox{\linewidth}{!}{%
\begin{tabular}{lrrrrrrrrrrrr}
\toprule
& \multicolumn{6}{c}{IDN Reg.} & \multicolumn{6}{c}{No Reg.} \\
\cmidrule(lr){2-7}\cmidrule(lr){8-13}
Layer & $\sigma_{\max}$ & $\|J\|_F$ & amp. & $10^3|\epsilon_l|$ & $10^3 C_\epsilon$ & $\Delta_g$ & $\sigma_{\max}$ & $\|J\|_F$ & amp. & $10^3|\epsilon_l|$ & $10^3 C_\epsilon$ & $\Delta_g$ \\
\midrule
25 & 1.66 & 2.40 & 2.92 & 0.271 & 0.271 & 0.056 & 18.25 & 24.75 & 16.57 & 20.463 & 20.463 & 0.159 \\
27 & 1.75 & 2.54 & 2.79 & 0.458 & 0.827 & 0.142 & 21.93 & 27.01 & 14.91 & 46.101 & 61.991 & 0.272 \\
29 & 1.69 & 2.45 & 2.55 & 1.453 & 2.379 & 0.277 & 19.87 & 25.66 & 12.29 & 66.422 & 123.962 & 0.436 \\
31 & 1.97 & 2.84 & 2.73 & 1.250 & 3.895 & 0.269 & 30.26 & 34.44 & 13.43 & 239.148 & 351.378 & 0.717 \\
\bottomrule
\end{tabular}%
}
\caption{Per-layer Jacobian, amplification, and substitution-error diagnostics on selected suffix layers of 0.5B standard models. The \(10^3|\epsilon_l|\) and \(10^3 C_\epsilon\) columns report relative substitution errors multiplied by 1000.}
\label{tab:jacobian_norms}
\label{tab:amplification}
\label{tab:variance_reduction}
\vspace{-1.5em}
\end{table*}

\begin{table}[t]
\centering
\resizebox{\linewidth}{!}{%
\begin{tabular}{lrrrrrrrrr}
\toprule
& \multicolumn{4}{c}{Sequential} & \multicolumn{5}{c}{IDN ($K=1$)} \\
\cmidrule(lr){2-5}\cmidrule(lr){6-10}
$N$ & Forward & Reversed & Best shuffle & Shuffle std. & Forward & Reversed & Best shuffle & No correction & Shuffle std. \\
\midrule
8 & \textbf{16.8} & 16.9 & 16.8 & 0.05 & \textbf{17.2} & 37.4 & 17.6 & 37.4 & 6.63 \\
12 & \textbf{16.8} & 17.2 & 16.9 & 0.06 & 17.9 & 67.2 & \textbf{17.8} & 67.2 & 25.62 \\
16 & \textbf{16.8} & 316.4 & 17.0 & 2793.71 & \textbf{18.0} & div. & 30.0 & div. & 565.01 \\
\bottomrule
\end{tabular}%
}
\caption{Correction-ordering summary for the 0.5B IDN Reg. model at sequence length 2048. Best shuffle is the lowest PPL among random shuffles; shuffle std. is computed over finite random-shuffle PPL values; \texttt{div.} denotes divergence. All IDN runs use \texttt{h0} initialization.}
\label{tab:ordering_summary}
\end{table}

\subsection{Inference effect}
\label{sec:inference_effect}

\noindent\textbf{Correction ordering.}
\cref{tab:ordering_summary} tests whether correction order matters on the IDN Reg. model; full tables are in \cref{app:ordering}. Sequentially executing the permuted suffix is almost invariant to layer order up to $N=12$, with very low shuffle variance, suggesting that IDN regularization makes these suffix layers effectively parallelizable. SNLP with IDN correction is more order-sensitive: the best $K=1$ PPL in the summary is almost always achieved by forward order, possibly because the correction recurrence preserves the causal direction of the original depth computation.

\noindent\textbf{Correction propagation.}
Layer-activation ablations in \cref{app:correction-propagation} confirm the propagation pattern from \cref{sec:method}: with correction, every active suffix layer can affect the final output in one iteration; without correction, influence moves only one layer per iteration, producing a staircase pattern.

\noindent\textbf{Variance reduction.}
For each suffix layer $l$, write $f_l(h)=x_{\mathrm{in},l}(h)+g_l(h)$, where $g_l$ is the non-residual attention/MLP update. Let $h_S$ be the clean prefix state and $h_l^{\mathrm{seq}}$ the sequential input to layer $l$. We define the substitution error $\epsilon_l=g_l(h_l^{\mathrm{seq}})-g_l(h_S)$ and report:
\begin{equation}
    |\epsilon_l|
    =
    \frac{
    \mathbb{E}_{b}\left[\|\epsilon_{l,b}\|_2\right]
    }{
    \mathbb{E}_{b}\left[\|h_{S,b}\|_2\right] + \epsilon
    },
    \qquad
    C_\epsilon(l)
    =
    \frac{
    \mathbb{E}_{b}\left[
    \left\|\sum_{r=S+1}^{l}\epsilon_{r,b}\right\|_2
    \right]
    }{
    \mathbb{E}_{b}\left[\|h_{S,b}\|_2\right] + \epsilon
    },
    \qquad
    \Delta_g(l)
    =
    \frac{
    \mathbb{E}_{b}\left[
    \left\|g_l(h_{l,b}^{\mathrm{seq}})-g_l(h_{S,b})\right\|_2
    \right]
    }{
    \mathbb{E}_{b}\left[\left\|g_l(h_{S,b})\right\|_2\right] + \epsilon
    }.
\end{equation}
Here $|\epsilon_l|$ measures the per-layer substitution error relative to the prefix-state scale, $C_\epsilon(l)$ measures how these errors accumulate through the suffix, and $\Delta_g(l)$ measures sensitivity relative to the layer update magnitude. \cref{tab:variance_reduction} shows that IDN Reg. keeps relative per-layer error in the 0.03\%--0.15\% range of $\|h_S\|$, while No Reg. ranges from 2\% to 24\%; IDN Reg. also has lower $\Delta_g$ than No Reg. at every reported layer.

\begin{wraptable}{r}{0.4\linewidth}
\vspace{-1.75em}
\centering
\footnotesize
\begin{tabular}{rrrr}
\toprule
$N$ & Seq layers & IDN Reg. & No Reg. \\
\midrule
0 & 32 & 16.82 & 15.21 \\
8 & 24 & 50.61 & 60.29 \\
12 & 20 & 156.67 & 285.32 \\
16 & 16 & 1785.02 & 4295.03 \\
\bottomrule
\end{tabular}
\caption{Early-exit PPL ($K=0$) for the 0.5B models at sequence length 2048. $N=0$ is ordinary sequential inference.
% ; larger $N$ skips the final $N$ layers.
}
\label{tab:early_exit_h5v3}
\vspace{-5em}
\end{wraptable}

\noindent\textbf{What $J\approx I$ really means.}
Input-invariant suffix layers are not droppable. For the highly parallel IDN Reg. checkpoint used in our main 0.5B results and most ablations, skipping 8--16 suffix layers greatly increases PPL (\cref{tab:early_exit_h5v3}). Thus $J\approx I$ means that the non-residual features are nearly input-invariant, not unnecessary.

\section{Conclusion}

We introduced Structured Newton Layer Parallelism, a training and inference framework for relaxing the strict layerwise dependency in Transformer inference. SNLP treats the hidden-state trace across depth as a residual equation, but replaces exact layer Jacobians with cheap structured surrogates induced by the architecture or training objective. In residual Transformers this yields IDN, while mHC-style models use HCN with a learned matrix surrogate.
These structured corrections make layer-parallel decoding practical enough to expose a useful speed-quality frontier: on 0.5B Nanochat models, SNLP reaches up to $2.58\times$ wall-clock speedup, and a less aggressive configuration reaches $1.40\times$ speedup without increasing PPL. The finite-iteration SNLP computation is not simply an exact numerical approximation to sequential execution; IDN and HCN introduce solver-induced inference bias, and the useful behavior comes from using this biased finite-iteration computation to trade quality for latency rather than forcing exact recovery of the sequential trace. This also makes SNLP useful beyond standalone PPL evaluation: SNLP-forward SFT can preserve downstream task accuracy, and SNLP drafters can accelerate self-speculative decoding while a sequential verifier preserves the final output.

\paragraph{Limitations.}
SNLP exposes a tradeoff rather than a guarantee of better quality, and the strongest wall-clock gains in our current implementation appear at the 0.5B scale. Larger models are not out of reach: the 3B IDN-regularized model already shows a $1.20\times$ speedup point with 3.9\% PPL increase, but wider sequential blocks make the systems problem harder. Early off-the-shelf experiments also show that making a pretrained model SNLP-compatible through lightweight finetuning is non-trivial, motivating stronger post-training adaptation.

\paragraph{Future directions.}
Promising directions include SNLP finetuning for off-the-shelf models, more systematic post-training and SFT studies, better kernels and algorithm--hardware co-design for larger models, and implementation work that could make the DiagN path practical despite its Jacobian-estimation cost. SNLP may also compose with other iterative decoding methods such as Jacobi decoding, apply to architectures beyond autoregressive Transformers such as diffusion language models, and help expose layer-parallel structure for settings with unusual execution constraints, including Homomorphic Encryption (HE).

% \paragraph{Broader impact.}
% More efficient layer-parallel inference could reduce serving cost and energy use, but the same efficiency gains may also make language-model deployment cheaper for harmful applications; our work does not change model capabilities or safety properties directly.



% \newpage
\bibliographystyle{abbrv}
\bibliography{ref}

\newpage
\appendix

\section*{\LARGE Appendix}
\markboth{Appendix}{Appendix}

\startcontents[appendix]

\begingroup
\contentsmargin{0pt}

\titlecontents{section}
  [1.4em]
  {}
  {\makebox[2.0em][l]{\contentslabel{1.6em}}}
  {}
  {\titlerule*[0.5pc]{.}\contentspage}

\titlecontents{subsection}
  [3.2em]
  {}
  {\makebox[3.0em][l]{\contentslabel{2.6em}}}
  {}
  {\titlerule*[0.5pc]{.}\contentspage}

\printcontents[appendix]{}{1}{\setcounter{tocdepth}{2}}
\endgroup
\vspace{1em}

\section{Note on Changes from the Previous Version}
\label{app:version-note}

The previous arXiv version reported main observations from PPL evaluation at sequence length 128, emphasizing short-context behavior. This version re-runs the main experiments at sequence length 2048, matching the training sequence length and giving a more representative long-context evaluation. As a result, the central interpretation changes: SNLP-aware training and IDN/HCN inference are framed as exposing a practical speed-quality frontier, rather than as a general mechanism for improving sequential PPL or beating sequential inference quality. This version also adds downstream task experiments with SFT and self-speculative decoding experiments with SNLP drafters and sequential verification.

\section{Algorithm}
\label{app:algo}

The main SNLP update is given in \Eqref{eq:snlp-update}. \Cref{alg:sequential-forward} shows the standard layer-by-layer execution, while \cref{alg:idn-batched} shows the batched structured-correction implementation used by the \texttt{NxF1-h0} configurations. For HCN, $A_l$ acts only on the HC/mHC stream dimension rather than the full hidden dimension.

\begin{algorithm}[h]
\caption{Standard Sequential Layer Execution}
\label{alg:sequential-forward}
\begin{algorithmic}[1]
\REQUIRE input tokens $x$, Transformer blocks $f_1,\ldots,f_L$
\STATE $h_0 \leftarrow \operatorname{Embed}(x)$
\FOR{$l=1,\ldots,L$}
    \STATE $h_l \leftarrow f_l(h_{l-1})$
\ENDFOR
\STATE \textbf{return} $\operatorname{LMHead}(h_L)$
\end{algorithmic}
\end{algorithm}

\begin{algorithm}[h]
\caption{IDN/HCN Batched SNLP Inference with \texttt{h0} Initialization}
\label{alg:idn-batched}
\begin{algorithmic}[1]
\REQUIRE input tokens $x$, prefix length $S$, suffix length $N=L-S$, iterations $K$
\STATE $h_0 \leftarrow \operatorname{Embed}(x)$
\FOR{$l=1,\ldots,S$}
    \STATE $h_l \leftarrow f_l(h_{l-1})$
\ENDFOR
\STATE $h_S^{(0)} \leftarrow h_S$; \quad $h_{S+j}^{(0)} \leftarrow h_S$ for $j=1,\ldots,N$
\FOR{$k=0,\ldots,K-1$}
    \STATE \textbf{parallel for} $j=1,\ldots,N$: \quad $\tilde h_{S+j}^{(k)} \leftarrow f_{S+j}\!\left(h_{S+j-1}^{(k)}\right)$
    \STATE $h_S^{(k+1)} \leftarrow h_S$
    \FOR{$j=1,\ldots,N$}
        \STATE $h_{S+j}^{(k+1)} \leftarrow \tilde h_{S+j}^{(k)} + A_{S+j}^{(k)}\!\left(h_{S+j-1}^{(k+1)} - h_{S+j-1}^{(k)}\right)$ ~~~~~~~~~~~~\COMMENT{$A=I$ for IDN; \Eqref{eq:hcn-A} for HCN.}
    \ENDFOR
\ENDFOR
\STATE \textbf{return} $\operatorname{LMHead}(h_L^{(K)})$
\end{algorithmic}
\end{algorithm}

\begin{algorithm}[h]
\caption{DiagN Inference with Associative Scan}
\label{alg:diagn-scan}
\begin{algorithmic}[1]
\REQUIRE input tokens $x$, prefix length $S$, suffix length $N=L-S$, iterations $K$
\STATE Run the sequential prefix to obtain $h_S$; initialize $h_{S+j}^{(0)}\leftarrow h_S$ for $j=1,\ldots,N$
\FOR{$k=0,\ldots,K-1$}
    \STATE \textbf{parallel for} $j=1,\ldots,N$: compute $\tilde h_{S+j}^{(k)} \leftarrow f_{S+j}(h_{S+j-1}^{(k)})$
    \STATE \textbf{parallel for} $j=1,\ldots,N$: estimate $a_j \leftarrow \operatorname{diag}\!\left(J_{S+j}^{(k)}\right)$ by Hutchinson FD/JVP/VJP probes
    \STATE $b_j \leftarrow \tilde h_{S+j}^{(k)} - a_j \odot h_{S+j-1}^{(k)}$ for $j=1,\ldots,N$
    \STATE Solve the affine recurrence $h_{S+j}^{(k+1)} = a_j\odot h_{S+j-1}^{(k+1)} + b_j$ with an associative prefix scan.
\ENDFOR
\STATE \textbf{return} $\operatorname{LMHead}(h_L^{(K)})$
\end{algorithmic}
\end{algorithm}

In \cref{alg:diagn-scan}, FD, JVP, and VJP change only how the diagonal estimator is obtained; once $a_j=\operatorname{diag}(J_{S+j}^{(k)})$ is available, all variants use the same affine scan recurrence.

\section{Analysis Details}
\label{app:analysis}

\subsection{Variance Reduction Derivation}
\label{app:variance}

Consider a residual suffix beginning at prefix state $h_S$, with
\begin{equation}
    f_l(h)=h+g_l(h), \qquad l=S+1,\ldots,L .
\end{equation}
The sequential suffix computes
\begin{equation}
    h_L^{\mathrm{seq}}
    =
    h_S + \sum_{l=S+1}^{L} g_l(h_{l-1}^{\mathrm{seq}}),
\end{equation}
where each branch $g_l$ is evaluated at a different accumulated hidden state. In contrast, one-step IDN initialized from $h_S$ computes
\begin{equation}
    h_L^{\mathrm{idn}}
    =
    h_S + \sum_{l=S+1}^{L} g_l(h_S).
\end{equation}
Thus the difference is entirely due to evaluation points:
\begin{equation}
    h_L^{\mathrm{seq}} - h_L^{\mathrm{idn}}
    =
    \sum_{l=S+1}^{L}
    \left[g_l(h_{l-1}^{\mathrm{seq}})-g_l(h_S)\right].
\end{equation}
First-order expansion around $h_S$ gives
\begin{equation}
    g_l(h_{l-1}^{\mathrm{seq}})-g_l(h_S)
    \approx
    J_{g_l}(h_S)\left(h_{l-1}^{\mathrm{seq}}-h_S\right)
    =
    J_{g_l}(h_S)
    \sum_{r=S+1}^{l-1} g_r(h_{r-1}^{\mathrm{seq}}).
\end{equation}
Sequential execution therefore couples every later layer deviation to all earlier deviations through the branch Jacobians. If the branch contributions have covariance scale $\sigma^2$ and $\|J_{g_l}\|\le \rho$ over the suffix, this term contributes a variance scale of order $\rho^2\sum_l(l-S)\sigma^2$ under a first-order independence approximation. IDN removes this compounding term by evaluating all branches at the same prefix state. The price is bias:
\begin{equation}
    \operatorname{Bias}_{\mathrm{idn}}
    =
    \sum_{l=S+1}^{L}
    \left[g_l(h_S)-g_l(h_{l-1}^{\mathrm{seq}})\right].
\end{equation}
SNLP-aware training directly reduces this bias by making the branch functions less sensitive over the suffix trajectory. When the induced bias is small, the variance reduction from avoiding chain-wise error compounding helps explain why finite-iteration SNLP can remain close enough to the sequential path to support a useful speed-quality tradeoff.

\subsection{Connection to Hyper-Connections and mHC}
\label{app:hc}

Hyper-Connections expand the residual stream into $M$ streams and mix them through learned matrices~\cite{zhu2025hyper}. A simplified HC sublayer can be written as
\begin{equation}
    x_{l+1}
    =
    H^{\mathrm{res}}_l x_l
    +
    H^{\mathrm{post}}_l
    F_l\!\left(H^{\mathrm{pre}}_l x_l\right),
\end{equation}
where the $H$ matrices act on stream dimension and $F_l$ is the nonlinear branch. For an mHC Transformer block, the attention and MLP wrappers each have their own residual mixing:
\begin{align}
    x'_l
    &=
    H^{\mathrm{res}}_{\mathrm{attn},l} x_l
    +
    H^{\mathrm{post}}_{\mathrm{attn},l}
    \operatorname{Attn}_l\!\left(H^{\mathrm{pre}}_{\mathrm{attn},l}x_l\right),\\
    x_{l+1}
    &=
    H^{\mathrm{res}}_{\mathrm{mlp},l} x'_l
    +
    H^{\mathrm{post}}_{\mathrm{mlp},l}
    \operatorname{MLP}_l\!\left(H^{\mathrm{pre}}_{\mathrm{mlp},l}x'_l\right).
\end{align}
The exact sublayer Jacobians contain both the residual mixing and the nonlinear branch sensitivity:
\begin{align}
    J_{\mathrm{attn},l}
    &=
    H^{\mathrm{res}}_{\mathrm{attn},l}
    +
    H^{\mathrm{post}}_{\mathrm{attn},l}
    J_{\operatorname{Attn}_l}
    H^{\mathrm{pre}}_{\mathrm{attn},l},\\
    J_{\mathrm{mlp},l}
    &=
    H^{\mathrm{res}}_{\mathrm{mlp},l}
    +
    H^{\mathrm{post}}_{\mathrm{mlp},l}
    J_{\operatorname{MLP}_l}
    H^{\mathrm{pre}}_{\mathrm{mlp},l}.
\end{align}
If training makes the nonlinear branches locally input-invariant, the branch Jacobian terms become small and the block Jacobian is approximated by
\begin{equation}
    J_{\mathrm{block},l}
    \approx
    H^{\mathrm{res}}_{\mathrm{mlp},l}
    H^{\mathrm{res}}_{\mathrm{attn},l}.
\end{equation}
This is exactly the HCN surrogate used in \Eqref{eq:snlp-update}. It is available from the architecture and acts only on stream dimension, avoiding full hidden-state Jacobian estimation.

IDN is the degenerate residual case. For $f_l(x)=x+g_l(x)$, the known residual Jacobian is $I$. IDN uses $A_l=I$ and trains the branch sensitivity $J_{g_l}$ to be small enough that $J_{f_l}=I+J_{g_l}\approx I$. In this sense, IDN and HCN follow the same template: keep the cheap architecture-induced residual transition, and train the nonlinear branch to be compatible with it.

\subsection{Fused Cross-Layer Coupling}
\label{app:fused-coupling}

Layer fusion combines several parallel layers that read the same input $h$ into one wide layer. For a chunk $\mathcal C$, the fused attention computes the branch projections for all $l\in\mathcal C$ and sums the output projections:
\begin{equation}
    a_{\mathcal C}(h)
    =
    \sum_{l\in\mathcal C}
    O_l\,\operatorname{Attn}_l(Q_lh,K_lh,V_lh).
\end{equation}
The fused MLP then receives the shared post-attention state
\begin{equation}
    u_{\mathcal C}(h)=h+a_{\mathcal C}(h),
\end{equation}
and applies the concatenated expansion and down-projection weights, equivalently summing the per-layer MLP branches:
\begin{equation}
    m_{\mathcal C}(h)
    =
    \sum_{l\in\mathcal C}
    P_l\,\phi\!\left(W_l\,\operatorname{Norm}(u_{\mathcal C}(h))\right),
\end{equation}
where $\phi$ is the MLP nonlinearity. The fused chunk output is
\begin{equation}
    F_{\mathcal C}^{\mathrm{fused}}(h)
    =
    u_{\mathcal C}(h)+m_{\mathcal C}(h).
\end{equation}
This is not identical to independently evaluating and summing each layer branch, because every MLP branch sees the aggregate attention state $h+\sum_{r\in\mathcal C}a_r(h)$ rather than only its own $h+a_l(h)$. The cross-layer coupling term is
\begin{equation}
    \sum_{l\in\mathcal C}
    P_l
    \left[
    \phi\!\left(W_l\operatorname{Norm}\!\left(h+\sum_{r\in\mathcal C}a_r(h)\right)\right)
    -
    \phi\!\left(W_l\operatorname{Norm}\!\left(h+a_l(h)\right)\right)
    \right].
\end{equation}
This term captures the change induced by evaluating every MLP branch on the aggregate post-attention state rather than on its own per-layer post-attention state. It is generally nonzero: layer normalization, nonlinear MLP activations, and non-canceling projections can all make a branch react to attention evidence produced by other layers in the same chunk. This implicit cross-layer coupling connects fused SNLP to the broader observation that architectures can expose useful depth mixing, as in HC/mHC and AttnRes~\cite{zhu2025hyper,xie2025mhc,chen2026attention}, while also explaining why overly aggressive fusion can change model behavior.

\section{Training Configuration and Ablations}
\label{app:training-config}

\cref{tab:vanilla_jacobi_spectral} summarizes an early training-side ablation that motivated SNLP-aware regularization. Vanilla Jacobi, which initializes every layer state from the prefix state and repeatedly applies $h_l^{(k+1)}=f_l(h_{l-1}^{(k)})$, diverged on trained models: measured layer Jacobian norms were far from contractive. Spectral regularization reduced this norm substantially, but it only modestly reduced the number of Newton iterations and did not address the dominant cost of Jacobian estimation. This motivated the IDN/HCN direction: instead of making exact Newton cheaper, train the model so a cheap structured correction is useful.

\begin{table}[t]
\centering
\small
\resizebox{1\linewidth}{!}{%
\begin{tabular}{llll}
\toprule
Attempt & Training target & Observation & Outcome \\
\midrule
Vanilla Jacobi & None & $\sigma_{\max}\!\approx\!44$; fixed-point iteration diverges & Failed \\
Spectral Reg. & Penalize $\operatorname{ReLU}(\hat\sigma_l-1)$ & $\sigma_{\max}$ reduced to $\approx\!2.3$, but iterations only drop about 20\% & Insufficient \\
IDN/HCN Reg. & Match parallel states to sequential & Removes JVP from correction and makes cheap structured updates effective & Used \\
\bottomrule
\end{tabular}%
}
\caption{Summary of early training-side attempts. Spectral regularization improved contractiveness but targeted Jacobian magnitude rather than the practical bottleneck: computing or estimating Jacobian corrections during inference.}
\label{tab:vanilla_jacobi_spectral}
\end{table}

\cref{tab:ckpt_selection} lists the training settings used for the trained-from-scratch models in \cref{tab:main_results}. \cref{tab:idn_reg_ablation} shows that the best loss configuration depends on architecture and scale: the 3B model prefers a smaller parallel suffix, the 0.5B standard model is sensitive to both $\lambda$ and stride, and the no-x0/VE model works best with stride 6. Detaching the MSE target hurts IDN PPL, while HCN on the mHC model uses a detached target by default.

\begin{table}[t]
\centering
\small
% \resizebox{0.95\linewidth}{!}{%
\begin{tabular}{llrrrrl}
\toprule
Model & Regularization & Steps & $N$ & $\lambda$ & Stride & Detach \\
\midrule
\multirow{2}{*}{Nanochat-3B standard}
& None & 9600 & -- & -- & -- & -- \\
& IDN & 9600 & 8 & 0.5 & 0 & \xmark \\
\midrule
\multirow{2}{*}{Nanochat-0.5B standard}
& None & 4800 & -- & -- & -- & -- \\
& IDN & 4800 & 24 & 0.0625 & 3 & \xmark \\
\midrule
\multirow{2}{*}{Nanochat-0.5B w/o x0ve}
& None & 4800 & -- & -- & -- & -- \\
& IDN & 4800 & 24 & 0.5 & 6 & \xmark \\
\midrule
\multirow{2}{*}{Nanochat-0.5B-mHC}
& None & 4800 & -- & -- & -- & -- \\
& HCN & 4800 & 24 & 0.5 & 0 & \cmark \\
\bottomrule
\end{tabular}%
% }
\caption{Training and SNLP-aware regularization settings for models used in the main results. Detach indicates whether the sequential target in \Eqref{eq:snlp-training-loss} is detached.}
\label{tab:ckpt_selection}
\end{table}

\begin{table}[t]
\centering
\small
\resizebox{0.95\linewidth}{!}{%
\begin{tabular}{llrrrrlrr}
\toprule
Model & Regularization & Steps & $N$ & $\lambda$ & Stride & Detach & PPL & $\Delta$PPL \\
\midrule
\multirow{10}{*}{\shortstack[l]{Nanochat-3B standard\\{[No Reg. PPL: 10.10]}}}
& \cellcolor{lightblue}IDN & \cellcolor{lightblue}9600 & \cellcolor{lightblue}8 & \cellcolor{lightblue}0.5 & \cellcolor{lightblue}0 & \cellcolor{lightblue}\xmark & \cellcolor{lightblue}10.07 & \cellcolor{lightblue}\good{-0.3\%} \\
& IDN & 9600 & 24 & 0.1 & 0 & \xmark & 10.12 & \comp{+0.2\%} \\
& IDN & 9600 & 24 & 0.1 & 3 & \xmark & 10.59 & \bad{+4.9\%} \\
& IDN & 9600 & 24 & 0.1 & 6 & \xmark & 10.36 & \bad{+2.6\%} \\
& IDN & 9600 & 24 & 0.5 & 0 & \xmark & 11.41 & \bad{+13.0\%} \\
& DiagN & 9600 & 8 & 0.05 & 3 & \cmark & 10.23 & \comp{+1.3\%} \\
& DiagN & 9600 & 8 & 0.1 & 3 & \cmark & 10.26 & \comp{+1.6\%} \\
& DiagN & 9600 & 24 & 0.5 & 3 & \cmark & 10.31 & \comp{+2.1\%} \\
& DiagN & 9600 & 24 & 0.1 & 3 & \cmark & 10.08 & \good{-0.2\%} \\
& DiagN & 9600 & 8 & 0.5 & 3 & \cmark & 10.43 & \bad{+3.3\%} \\
\midrule
\multirow{13}{*}{\shortstack[l]{Nanochat-0.5B standard\\{[No Reg. PPL: 15.21]}}}
& IDN & 4800 & 24 & 0.5 & 0 & \xmark & 15.23 & \comp{+0.1\%} \\
& IDN & 4800 & 24 & 0.1 & 0 & \xmark & 15.34 & \comp{+0.9\%} \\
& \cellcolor{lightblue}IDN & \cellcolor{lightblue}4800 & \cellcolor{lightblue}24 & \cellcolor{lightblue}0.0625 & \cellcolor{lightblue}3 & \cellcolor{lightblue}\xmark & \cellcolor{lightblue}15.36 & \cellcolor{lightblue}\comp{+1.0\%} \\
& IDN & 4800 & 24 & 0.1 & 3 & \xmark & 15.43 & \comp{+1.5\%} \\
& IDN & 4800 & 24 & 0.5 & 0 & \cmark & 16.46 & \bad{+8.3\%} \\
& IDN & 4800 & 24 & 0.5 & 3 & \xmark & 16.81 & \bad{+10.5\%} \\
& IDN & 4800 & 24 & 0.5 & 2 & \xmark & 21.36 & \bad{+40.5\%} \\
& IDN & 4800 & 24 & 0.5 & 3 & \cmark & NaN & -- \\
& IDN & 2040 & 24 & 0.5 & 0 & \xmark & 17.54 & \bad{+15.3\%} \\
& DiagN & 4800 & 24 & 0.1 & 3 & \cmark & 15.18 & \good{-0.2\%} \\
& DiagN & 4800 & 24 & 0.5 & 3 & \cmark & 17.13 & \bad{+12.6\%} \\
& DiagN & 2040 & 24 & 0.1 & 3 & \cmark & 17.08 & \bad{+12.3\%} \\
& DiagN & 2040 & 24 & 0.5 & 3 & \cmark & 26.70 & \bad{+75.5\%} \\
\midrule
\multirow{3}{*}{\shortstack[l]{Nanochat-0.5B w/o x0ve\\{[No Reg. PPL: 17.65]}}}
& \cellcolor{lightblue}IDN & \cellcolor{lightblue}4800 & \cellcolor{lightblue}24 & \cellcolor{lightblue}0.5 & \cellcolor{lightblue}6 & \cellcolor{lightblue}\xmark & \cellcolor{lightblue}17.57 & \cellcolor{lightblue}\good{-0.4\%} \\
& IDN & 4800 & 24 & 0.5 & 0 & \xmark & 17.63 & \good{-0.1\%} \\
& IDN & 4800 & 24 & 0.5 & 3 & \xmark & 20.33 & \bad{+15.2\%} \\
\midrule
\multirow{3}{*}{\shortstack[l]{Nanochat-0.5B-mHC\\{[No Reg. PPL: 15.16]}}}
& \cellcolor{lightblue}HCN & \cellcolor{lightblue}4800 & \cellcolor{lightblue}24 & \cellcolor{lightblue}0.5 & \cellcolor{lightblue}0 & \cellcolor{lightblue}\cmark & \cellcolor{lightblue}15.52 & \cellcolor{lightblue}\comp{+2.4\%} \\
& HCN & 4800 & 24 & 0.5 & 0 & \xmark & 15.91 & \bad{+4.9\%} \\
& HCN & 4800 & 24 & 0.5 & 3 & \cmark & 21.67 & \bad{+43.0\%} \\
\bottomrule
\end{tabular}%
}
\caption{Ablation of SNLP-aware regularization loss settings. Detach indicates whether the sequential target is detached. PPL is evaluated at sequence length 2048. $\Delta$PPL is relative to the corresponding No Reg. baseline shown under each model group. Selected main-result checkpoints are highlighted.}
\label{tab:idn_reg_ablation}
\end{table}

\section{Additional Inference Ablations}
\label{app:inference-ablations}

\noindent\textbf{Checkpoint note.}
The correction-ordering and correction-propagation ablations in this appendix use the more parallelizable 0.5B IDN checkpoint from \cref{tab:idn_reg_ablation} (\(N=24\), \(\lambda=0.5\), stride 3, no detach; PPL 16.81).

\subsection{Interpreting AR Match Rate}
\label{app:ar-match-rate}

\cref{tab:ar_match_rate_conversion} gives a rough conversion between the observed greedy autoregressive match rate and an effective local per-token agreement rate. For AR Match, each sample is evaluated over $T=32$ generated tokens. The reported AR match rate $\alpha$ is token-level agreement against the sequential baseline over $N$ samples $\alpha = \frac{\#\text{matched generated tokens}}{N T}$.
Because generation is autoregressive, a single mismatch changes the prefix for all later positions. Under a simple absorbing-divergence model, let $\beta$ be the probability of matching the next token conditioned on all previous generated tokens matching. Then token $i$ matches with probability $\beta^i$, so
\[
\alpha = \frac{1}{T}\sum_{i=1}^{T}\beta^i
= \frac{\beta(1-\beta^T)}{T(1-\beta)} .
\]
Thus an apparently modest AR match can still imply high local agreement before prefix divergence; for $T=32$, $\alpha=0.30$ corresponds to $\hat\beta\approx0.905$.

\begin{table}[t]
\centering
\begin{tabular}{rr}
\toprule
Observed AR match $\alpha$ & Implied conditional match $\hat\beta$ \\
\midrule
0.95 & 0.9969 \\
0.90 & 0.9935 \\
0.80 & 0.9859 \\
0.70 & 0.9767 \\
0.60 & 0.9654 \\
0.50 & 0.9510 \\
0.40 & 0.9320 \\
0.30 & 0.9048 \\
\bottomrule
\end{tabular}
\caption{Conversion from observed token-level AR match rate to effective conditional per-token agreement for $T=32$ generated tokens under an absorbing-divergence model.}
\label{tab:ar_match_rate_conversion}
\end{table}

\subsection{Coupling SNLP with Jacobi Decoding}
\label{app:sjd-coupling}

SNLP parallelizes over the layer axis, while Jacobi decoding (JD) parallelizes over future token positions by iteratively refining a block of draft tokens~\cite{santilli2023accelerating}. Speculative decoding and speculative Jacobi decoding (SJD) further combine draft proposals with verification~\cite{leviathan2023fast,chen2023accelerating,teng2025accelerating}. A natural extension is to couple the two axes and solve over a layer-token lattice
\[
    h_{\ell,t},
    \qquad
    \ell=0,\ldots,L,\quad t=1,\ldots,T .
\]
In principle, one could update hidden states and token guesses jointly,
\[
    h_{\ell,t}^{(r+1)}
    =
    \mathcal F_{\ell,t}\!\left(h^{(r)}, x^{(r)}\right),
    \qquad
    x_t^{(r+1)}
    =
    \operatorname{JDUpdate}\!\left(\operatorname{LMHead}(h_{L,t}^{(r+1)})\right),
\]
rather than using a nested loop,
\[
    x^{(r+1)}
    \leftarrow
    \operatorname{JDUpdate}\!\left(
    \operatorname{SNLPForward}(x^{(r)};K)
    \right).
\]
This view is appealing because a successful 2D solver could avoid paying a full inner SNLP solve for every token-level Jacobi iteration. In practice, the main difficulty is initialization. JD changes the draft tokens between iterations; reinitializing all layer states from the new prefix state \(h_0\) reduces to the nested baseline, while reusing hidden states from the previous JD iteration carries features computed for old draft tokens.

\Cref{tab:sjd_coupling} evaluates this design on a 0.5B IDN stride-0 checkpoint. The naive variants run a fresh SNLP forward inside each JD iteration. The h0-JD variants recompute the first parallel input but warm-start deeper parallel states from the previous JD iteration. Naive \(K=1\) already obtains 100\% token match across all tested configurations, leaving no inner-iteration gap for coupling to close. Increasing to \(K=2\) adds cost without improving match. By contrast, h0-JD degrades match to roughly 21--37\%, because stale hidden states encode features for old draft tokens and are propagated by the IDN correction. SJD-style variable-length acceptance makes direct hidden-state reuse even less straightforward, so the tested SJD path effectively reduces to naive composition. We therefore leave useful 2D layer-token coupling to future work; it likely requires a better transport or reinitialization rule when draft tokens change.

\begin{table}[t]
\centering
\scriptsize
\resizebox{\linewidth}{!}{%
\begin{tabular}{lrrrrrrrrrrrr}
\toprule
& \multicolumn{4}{c}{8xF1, $N=8$} & \multicolumn{4}{c}{12xF1, $N=12$} & \multicolumn{4}{c}{4xF3, $N=12$} \\
\cmidrule(lr){2-5}\cmidrule(lr){6-9}\cmidrule(lr){10-13}
Config & Match & Accept & JD iters & Fwd passes & Match & Accept & JD iters & Fwd passes & Match & Accept & JD iters & Fwd passes \\
\midrule
\textbf{naive $K=1$} & \textbf{100.0\%} & \textbf{1.04} & \textbf{1.96} & \textbf{30.9} & \textbf{100.0\%} & \textbf{1.13} & \textbf{1.91} & \textbf{28.8} & \textbf{100.0\%} & \textbf{1.05} & \textbf{1.96} & \textbf{30.4} \\
naive $K=2$ & 100.0\% & 1.02 & 1.98 & 31.5 & 100.0\% & 1.04 & 1.95 & 30.8 & 100.0\% & 1.06 & 1.93 & 30.2 \\
h0-JD $K=1$ & 25.0\% & 0.85 & 2.37 & 37.9 & 21.2\% & 0.70 & 2.88 & 45.9 & 22.8\% & 1.10 & 2.16 & 31.0 \\
h0-JD $K=2$ & 37.2\% & 1.00 & 2.07 & 31.9 & 27.8\% & 0.94 & 2.17 & 34.2 & 36.5\% & 1.03 & 2.01 & 31.1 \\
\bottomrule
\end{tabular}%
}
\caption{Coupling JD with SNLP on a 0.5B IDN stride-0 checkpoint, evaluated on 8 prompts with 32 generated tokens and lookahead window 5. Naive runs reinitialize SNLP hidden states in each JD iteration. \texttt{h0-JD} recomputes the first parallel input but reuses deeper parallel states from the previous JD iteration.}
\label{tab:sjd_coupling}
\end{table}

\subsection{ELK Tempering and Preheat Initialization}
\label{app:elk-preheat}

We also explored two auxiliary inference knobs that are not included in the main configuration search. The first is an ELK-style tempering of the Newton correction~\cite{gonzalez2024towards}, which scales the correction term without changing the number of block forwards and therefore should not affect speed. The second is preheat initialization: offline calibration fits a low-rank affine predictor for each layer output,
\[
    \widehat h_l(x_0)
    =
    (x_0 V_l) U_l^\top + b_l ,
\]
where the basis \(V_l\) is obtained from a truncated SVD of calibration embeddings and \(U_l,b_l\) are fit by linear regression to sequential hidden states. At inference time, \(\widehat h_l(x_0)\) initializes the parallel suffix. In the table below, preheat was calibrated on random tokens; later validation-set calibration did not consistently improve results.

\Cref{tab:elk_preheat} reports an auxiliary sequence-length-128 sweep and shows that ELK tempering can improve the PPL--iteration tradeoff for some configurations, but the effect is not uniform. Preheat is also inconsistent: it can be close to \texttt{h0}, but it can also be much worse. To reduce search space, our main experiments do not tune ELK or preheat; a more systematic search may expose different points on the speed-quality frontier.

\begin{table}[t]
\centering
\scriptsize
\resizebox{\linewidth}{!}{%
\begin{tabular}{llrrrrrrrrrrrr}
\toprule
& & \multicolumn{3}{c}{1xF12} & \multicolumn{3}{c}{2xF6} & \multicolumn{3}{c}{4xF3} & \multicolumn{3}{c}{12xF1} \\
\cmidrule(lr){3-5}\cmidrule(lr){6-8}\cmidrule(lr){9-11}\cmidrule(lr){12-14}
ELK & Init & h0 & fwd & preheat & h0 & fwd & preheat & h0 & fwd & preheat & h0 & fwd & preheat \\
\midrule
\multirow{1}{*}{0}
& PPL & 94.04 & 94.04 & 94.04 & 102.8 & 102.5 & 169.1 & 105.8 & 94.44 & 281.0 & 144.5 & 93.83 & 1218 \\
\midrule
\multirow{1}{*}{0.1}
& PPL & 94.04 & 94.04 & 94.04 & \good{97.94} & \good{93.90} & \good{124.7} & \good{91.75} & \good{82.43} & \good{124.2} & \bad{586.5} & \good{92.87} & \bad{81293} \\
\bottomrule
\end{tabular}%
}
\caption{ELK tempering and preheat initialization ablation for \(N=12\), \(K=1\), evaluated at sequence length 128 with sequential PPL 59.07. Configs follow the main chunk notation: \texttt{1xF12}, \texttt{2xF6}, \texttt{4xF3}, and \texttt{12xF1}. The \texttt{fwd} column corresponds to one-shot batched-forward initialization.}
\label{tab:elk_preheat}
\end{table}

\subsection{Off-the-Shelf Model Results}
\label{app:ots-models}

\cref{tab:ots_models} summarizes selected post-hoc SNLP configurations for off-the-shelf models from the L1 timing sweep. We report fast configurations under relaxed PPL thresholds and the lowest-PPL configuration. Matching sequential perplexity requires multiple Newton iterations and does not produce speedup, supporting the need for SNLP-aware training.

\begin{table}[t]
\centering
\scriptsize
\resizebox{0.85\linewidth}{!}{%
\begin{tabular}{lrlrrr}
\toprule
Model & Seq PPL & Config & $K$ & PPL & Speedup \\
\midrule
\multirow{2}{*}{Qwen2.5-0.5B} & \multirow{2}{*}{14.36} & \texttt{8xF1-fwd} & 4 & 15.55 (+8.2\%) & $0.74\times$ \\
& & \texttt{8xF1-fwd} & 8 & 14.36 (+0.0\%) & $0.64\times$ \\
\cmidrule(lr){1-6}
\multirow{3}{*}{TinyLlama-1.1B} & \multirow{3}{*}{8.03} & \texttt{8xF1-fwd} & 4 & 8.84 (+10.1\%) & $0.75\times$ \\
& & \texttt{12xF1-fwd} & 8 & 8.17 (+1.7\%) & $0.66\times$ \\
& & \texttt{8xF1-fwd} & 8 & 8.03 (+0.0\%) & $0.63\times$ \\
\bottomrule
\end{tabular}%
}
\caption{Selected post-hoc SNLP configurations on off-the-shelf models. Config names follow the chunk notation used in \cref{tab:main_results}; \texttt{8xF1-fwd} denotes 8 parallel single-layer chunks initialized from a one-shot parallel forward.}
\label{tab:ots_models}
\end{table}

\subsection{Diagonal-Jacobian Correction Results}
\label{app:diag-jacobian}

\begin{table*}[t]
\centering
\resizebox{\linewidth}{!}{%
\begin{tabular}{lllrrrrrrr}
\toprule
Model & Config & $K$ & PPL & $\Delta$PPL & Speedup & Top-1 & LogitSim & EmbSim & AR Match \\
\midrule

\multicolumn{10}{c}{\textbf{Nanochat-3B standard}} \\
\midrule
\multirow{2}{*}{\modelppl{No Reg.}{10.10}}
& {\small\texttt{8xF1-h0}} & 4 & 10.25 & \comp{+1.5\%} & \bad{$0.44\times$} & 92.2\% & 0.997 & 0.985 & 80.2\% \\
& {\small\texttt{8xF1-fwd}} & 4 & 10.20 & \comp{+1.0\%} & \bad{$0.42\times$} & 96.1\% & 0.999 & 0.990 & 88.9\% \\
\cmidrule(lr){1-10}

\multirow{2}{*}{\modelppl{IDN Reg.}{10.07 (\good{-0.3\%})}}
& {\small\texttt{8xF1-h0}} & 1 & 10.50 & \bad{+4.3\%} & \bad{$0.71\times$} & 85.8\% & 0.985 & 0.968 & 63.3\% \\
& {\small\texttt{8xF1-h0}} & 4 & 10.10 & \comp{+0.3\%} & \bad{$0.44\times$} & 98.1\% & 1.000 & 0.993 & 93.4\% \\

\midrule
\multicolumn{10}{c}{\textbf{Nanochat-0.5B standard}} \\
\midrule
\multirow{2}{*}{\modelppl{No Reg.}{15.21}}
& {\small\texttt{8xF1-fwd}} & 2 & 16.65 & \bad{+9.5\%} & \bad{$0.58\times$} & 82.4\% & 0.992 & 0.981 & 71.2\% \\
& {\small\texttt{8xF1-fwd}} & 4 & 15.45 & \comp{+1.6\%} & \bad{$0.44\times$} & 98.3\% & 1.000 & 0.998 & 97.3\% \\
\cmidrule(lr){1-10}

\multirow{2}{*}{\modelppl{IDN Reg.}{15.36 (\comp{+1.0\%})}}
& {\small\texttt{8xF1-h0}} & 1 & 16.30 & \bad{+6.1\%} & \bad{$0.72\times$} & 86.4\% & 0.991 & 0.984 & 79.9\% \\
& {\small\texttt{8xF1-h0}} & 2 & 15.53 & \comp{+1.1\%} & \bad{$0.60\times$} & 96.1\% & 0.999 & 0.994 & 92.2\% \\

\midrule
\multicolumn{10}{c}{\textbf{Nanochat-0.5B w/o x0ve}} \\
\midrule
\multirow{2}{*}{\modelppl{No Reg.}{17.65}}
& {\small\texttt{8xF1-h0}} & 2 & 19.35 & \bad{+9.6\%} & \bad{$0.62\times$} & 84.9\% & 0.997 & 0.985 & 78.3\% \\
& {\small\texttt{8xF1-h0}} & 4 & 17.76 & \comp{+0.6\%} & \bad{$0.48\times$} & 97.7\% & 1.000 & 0.999 & 97.9\% \\
\cmidrule(lr){1-10}

\multirow{2}{*}{\modelppl{IDN Reg.}{17.57 (\good{-0.5\%})}}
& {\small\texttt{20xF1-h0}} & 2 & 18.80 & \bad{+7.0\%} & \bad{$0.88\times$} & 77.2\% & 0.996 & 0.960 & 60.9\% \\
& {\small\texttt{8xF1-h0}} & 2 & 17.81 & \comp{+1.4\%} & \bad{$0.62\times$} & 90.9\% & 1.000 & 0.982 & 79.7\% \\

\midrule
\multicolumn{10}{c}{\textbf{Nanochat-0.5B-mHC}} \\
\midrule
\multirow{2}{*}{\modelppl{No Reg.}{15.16}}
& {\small\texttt{8xF1-h0}} & 4 & 16.03 & \bad{+5.7\%} & \bad{$0.46\times$} & 92.2\% & 0.996 & 0.992 & 87.6\% \\
& {\small\texttt{8xF1-fwd}} & 4 & 15.34 & \comp{+1.2\%} & \bad{$0.45\times$} & 96.9\% & 0.999 & 0.997 & 94.8\% \\
\cmidrule(lr){1-10}

\modelppl{HCN Reg.}{15.52 (\comp{+2.4\%})}
& {\small\texttt{8xF1-h0}} & 8 & 15.87 & \comp{+2.2\%} & \bad{$0.31\times$} & 92.6\% & 0.998 & 0.988 & 86.0\% \\
\bottomrule
\end{tabular}%
}
\caption{Main diagonal-Jacobian Newton correction results using 2048-token PPL evaluation. Brackets under model names give sequential PPL and, when shown, the change from the corresponding No Reg. baseline. Configs use the same chunk notation as \cref{tab:main_results}; all rows use DiagN correction.}
\label{tab:diag_jacobian_results}
\end{table*}

\cref{tab:diag_ops_bench} benchmarks the local cost of Jacobian-vector primitives used by the diagonal-correction sweep in \cref{tab:diag_jacobian_results}. Finite-difference (FD) estimates $Jv$ by evaluating $f(x+\epsilon v)-f(x)$, exact forward-mode JVP returns both $f(x)$ and $Jv$ in one call, and VJP uses reverse-mode autodiff to compute $J^\top v$. FD is efficient in our solver because $f(x)$ is already computed by the parallel block forward, so the finite difference only adds the extra $f(x+\epsilon v)$ evaluation, similar to the cost of forward initialization. However, FD often diverges in our sweep, plausibly because the correction stacks multiple approximations: replacing the full Jacobian by its diagonal, estimating that diagonal with a Hutchinson-style probe, and then using a noisy finite difference. JVP and VJP are more stable but expensive: \cref{tab:diag_ops_bench} shows roughly $3\times$ overhead for VJP and $6$--$8\times$ for JVP over a plain forward. Exact JVP also lacks fused-SDPA support, so the diagonal experiments use FD or VJP estimators.

In a short-context mHC diagnostic, no diagonal Newton configuration reaches the 10\% fast threshold relative to sequential PPL 88.67: the best diagonal result reaches 104.2 PPL, or +17.5\% above sequential. With the x0+VE mHC models in \cref{tab:diag_jacobian_results}, diagonal correction can match PPL but remains much slower. This failure mode is expected: mHC routing through $H_{\mathrm{res}}$, $H_{\mathrm{pre}}$, and $H_{\mathrm{post}}$ is asymmetric, so a VJP estimates $J^\top v$ rather than the required $Jv$ direction. The resulting diagonal approximation is poor, causing slow convergence or divergence.

\begin{table}[t]
\centering
\scriptsize
\resizebox{0.94\linewidth}{!}{%
\begin{tabular}{llrrrr}
\toprule
Model & $T$ & Forward (ms) & FD (ms) & JVP (ms) & VJP (ms) \\
\midrule
\multirow{5}{*}{Nanochat-3B standard}
& 1 & 0.458 & 0.816 ($1.78\times$) & 3.274 ($7.14\times$) & 1.346 ($2.94\times$) \\
& 16 & 0.459 & 0.959 ($2.09\times$) & 3.356 ($7.31\times$) & 1.369 ($2.98\times$) \\
& 32 & 0.458 & 0.926 ($2.02\times$) & 3.441 ($7.51\times$) & 1.379 ($3.01\times$) \\
& 64 & 0.463 & 0.952 ($2.05\times$) & 3.421 ($7.38\times$) & 1.355 ($2.92\times$) \\
& 128 & 0.454 & 1.006 ($2.22\times$) & 3.380 ($7.44\times$) & 1.345 ($2.96\times$) \\
\midrule
\multirow{5}{*}{Nanochat-0.5B standard}
& 1 & 0.418 & 0.784 ($1.88\times$) & 3.235 ($7.74\times$) & 1.330 ($3.18\times$) \\
& 16 & 0.420 & 0.833 ($1.98\times$) & 3.299 ($7.85\times$) & 1.292 ($3.08\times$) \\
& 32 & 0.411 & 0.845 ($2.06\times$) & 3.378 ($8.23\times$) & 1.339 ($3.26\times$) \\
& 64 & 0.424 & 0.872 ($2.05\times$) & 3.321 ($7.83\times$) & 1.309 ($3.08\times$) \\
& 128 & 0.421 & 0.832 ($1.98\times$) & 3.397 ($8.07\times$) & 1.334 ($3.17\times$) \\
\midrule
\multirow{5}{*}{Qwen2.5-0.5B}
& 1 & 0.522 & 1.059 ($2.03\times$) & 3.014 ($5.77\times$) & 1.753 ($3.35\times$) \\
& 16 & 0.545 & 1.091 ($2.00\times$) & 3.132 ($5.75\times$) & 1.855 ($3.40\times$) \\
& 32 & 0.545 & 1.090 ($2.00\times$) & 3.179 ($5.84\times$) & 1.875 ($3.44\times$) \\
& 64 & 0.542 & 1.071 ($1.97\times$) & 3.127 ($5.77\times$) & 1.843 ($3.40\times$) \\
& 128 & 0.537 & 1.080 ($2.01\times$) & 3.162 ($5.89\times$) & 1.852 ($3.45\times$) \\
\bottomrule
\end{tabular}%
}
\caption{Single-layer operation benchmark. Nanochat timings use layer 24, and Qwen2.5 timings use a standalone middle layer. Times are milliseconds over 100 warmup and 500 timed runs; parentheses report cost relative to a plain forward.}
\label{tab:diag_ops_bench}
\end{table}

\subsection{Pretrained TinyLlama Fine-Tuning}
\label{app:tinyllama-finetuning}

We briefly finetune TinyLlama-1.1B-Chat-v1.0~\cite{zhang2024tinyllama} with IDN regularization to test whether layer-parallel compatibility can be retrofitted onto an off-the-shelf pretrained model. TinyLlama is a standard LLaMA-style model with 22 layers, hidden dimension 2048, 32 attention heads, and 4 KV heads. We finetune on ClimbMix~\cite{diao2025climb} using AdamW with cosine learning-rate decay; \cref{tab:tinyllama_finetune_grid} summarizes the small grid.

\begin{table}[t]
\centering
\scriptsize
\resizebox{0.85\linewidth}{!}{%
\begin{tabular}{lrrrrr}
\toprule
Run & IDN weight & Target $N$ & LR & Steps & Final PPL \\
\midrule
baseline & 0 & -- & $2{\times}10^{-5}$ & 2000 & 17.82 \\
idn05 & 0.5 & 4,8,12 & $2{\times}10^{-5}$ & 2000 & 17.91 \\
idn2\_npar48 & 2.0 & 4,8 & $5{\times}10^{-5}$ & 5000 & 19.29 \\
idn5\_npar4 & 5.0 & 4 & $1{\times}10^{-4}$ & 5000 & 26.03 \\
idn10\_npar4 & 10.0 & 4 & $1{\times}10^{-4}$ & 5000 & 26.04 \\
idn5\_npar4812 & 5.0 & 4,8,12 & $5{\times}10^{-5}$ & 5000 & 19.53 \\
idn1\_long & 1.0 & 4,8 & $1{\times}10^{-5}$ & 4000* & $\approx 17.8$ \\
idn5\_long\_npar4 & 5.0 & 4 & $1{\times}10^{-5}$ & 4000* & $\approx 17.8$ \\
\bottomrule
\end{tabular}%
}
\caption{TinyLlama IDN finetuning grid. Final PPL uses sequence length 128 for quick early testing. Runs marked with * crashed before the planned 10000 steps; the checkpoint near 2000 steps was still usable for evaluation.}
\label{tab:tinyllama_finetune_grid}
\end{table}

These preliminary runs support the co-design interpretation: simple post-hoc finetuning is hard to use to change a pretrained model's layer-parallel behavior when the model was not pretrained with SNLP regularization. We leave stronger pretrained-model adaptation to future investigation.

\subsection{Correction Ordering}
\label{app:ordering}

\cref{tab:h5v3_n8,tab:h5v3_n12,tab:h5v3_n16} report the correction-ordering ablations used to construct the summary in \cref{tab:ordering_summary}. AR-ness is the average of local- and global-AR-ness scores as defined in DiffuCoder~\cite{gong2025diffucoder}. At $K=1$, several non-forward orders remain usable; after repeated correction at $K=4$, forward ordering consistently gives the best stable result, while most non-forward orders degrade or diverge.

\begin{table}[t]
\centering
\scriptsize
\resizebox{\linewidth}{!}{%
\begin{tabular}{lrrrrrrr}
\toprule
& & \multicolumn{3}{c}{IDN Reg.} & \multicolumn{3}{c}{No Reg.} \\
\cmidrule(lr){3-5}\cmidrule(lr){6-8}
Order & AR-ness & Seq-Perm & $K=1$ & $K=4$ & Seq-Perm & $K=1$ & $K=4$ \\
\midrule
forward & 1.000 & \textbf{16.8} & \textbf{17.2} & \textbf{16.9} & \textbf{15.2} & 1585.9 & \textbf{15.4} \\
shuffle\_15 & 0.384 & 16.8 & 24.4 & 281.6 & 98.1 & 32.1 & 496161.9 \\
shuffle\_0 & 0.339 & 16.8 & 19.5 & 46.4 & 2391.2 & 210.3 & 13103.0 \\
shuffle\_14 & 0.321 & 16.9 & 29.3 & 28.3 & 579.7 & 32.8 & 444232.8 \\
shuffle\_5 & 0.259 & 16.8 & 18.2 & 19.1 & 249.2 & 296.0 & div. \\
shuffle\_10 & 0.259 & 16.8 & 17.6 & 19.4 & 32.8 & 667.8 & div. \\
shuffle\_13 & 0.205 & 16.8 & 18.5 & 22.1 & 530.0 & 12736.3 & div. \\
shuffle\_9 & 0.196 & 16.8 & 20.2 & div. & 97.7 & 41.4 & 240814.7 \\
shuffle\_6 & 0.188 & 16.8 & 27.2 & div. & 39.5 & 28.1 & div. \\
shuffle\_11 & 0.188 & 16.9 & 29.3 & 32.0 & 654.4 & 32.8 & 768834.1 \\
shuffle\_12 & 0.188 & 16.8 & 23.7 & 44.2 & 40.7 & 28.7 & 139401.7 \\
shuffle\_2 & 0.134 & 16.8 & 18.2 & 41.5 & 91.4 & 29796.1 & 92160.6 \\
shuffle\_1 & 0.125 & 16.9 & 41.3 & 54.1 & 1028.0 & 32.9 & 241407.0 \\
shuffle\_3 & 0.125 & 16.9 & 22.9 & 32.3 & 48.1 & \textbf{27.2} & div. \\
shuffle\_4 & 0.125 & 16.8 & 28.5 & div. & 359.9 & 34.3 & 9585.0 \\
shuffle\_7 & 0.125 & 16.9 & 36.3 & 22.7 & 6281.4 & 33.9 & 185013.0 \\
shuffle\_8 & 0.125 & 16.9 & 22.3 & 50.7 & 156.8 & 30.5 & div. \\
reversed & 0.062 & 16.9 & 37.4 & 41.7 & 24622.8 & 37.3 & div. \\
no correction & -- & -- & 37.4 & 19.7 & -- & 37.3 & 20.0 \\
\bottomrule
\end{tabular}%
}
\caption{Correction ordering ablation for $N=8$ on the IDN Reg. checkpoint and No Reg. baseline, evaluated at sequence length 2048.}
\label{tab:h5v3_n8}
\end{table}

\begin{table}[t]
\centering
\scriptsize
\resizebox{\linewidth}{!}{%
\begin{tabular}{lrrrrrrr}
\toprule
& & \multicolumn{3}{c}{IDN Reg.} & \multicolumn{3}{c}{No Reg.} \\
\cmidrule(lr){3-5}\cmidrule(lr){6-8}
Order & AR-ness & Seq-Perm & $K=1$ & $K=4$ & Seq-Perm & $K=1$ & $K=4$ \\
\midrule
forward & 1.000 & \textbf{16.8} & 17.9 & \textbf{17.0} & \textbf{15.2} & 283.7 & \textbf{18.2} \\
shuffle\_7 & 0.390 & 17.0 & 31.7 & 37.4 & 15111.5 & 177.7 & 610793.2 \\
shuffle\_10 & 0.341 & 17.0 & 117.4 & 1210.0 & 90291.2 & 184.9 & div. \\
shuffle\_13 & 0.299 & 16.9 & 17.9 & 141.2 & 94.5 & 283.6 & div. \\
shuffle\_0 & 0.261 & 16.9 & 32.1 & div. & 11287.2 & 567.3 & div. \\
shuffle\_6 & 0.254 & 16.9 & 20.8 & div. & 135.9 & 95.5 & div. \\
shuffle\_9 & 0.212 & 16.9 & 23.7 & div. & 1259.9 & 81.3 & div. \\
shuffle\_5 & 0.170 & 16.9 & \textbf{17.8} & 23.2 & 82.0 & 283.6 & div. \\
shuffle\_11 & 0.167 & 16.9 & 28.4 & 441.4 & 10006.0 & \textbf{71.5} & div. \\
shuffle\_3 & 0.125 & 17.1 & 67.2 & 25.9 & 82494.6 & 219.7 & 247566.1 \\
shuffle\_4 & 0.125 & 16.9 & 67.2 & 27.6 & 1452.9 & 219.7 & div. \\
shuffle\_12 & 0.125 & 17.0 & 52.0 & div. & 3809.4 & 1286.6 & div. \\
shuffle\_1 & 0.087 & 16.9 & 19.2 & 98879.6 & 491.6 & 1105.8 & div. \\
shuffle\_2 & 0.083 & 17.0 & 28.7 & div. & 50156.7 & 346.7 & div. \\
shuffle\_14 & 0.083 & 17.0 & 35.4 & div. & 15778.7 & 72.6 & div. \\
shuffle\_8 & 0.042 & 17.0 & 21.9 & div. & 2751.0 & 1182.6 & div. \\
shuffle\_15 & 0.042 & 16.9 & 27.6 & 808.6 & 16138.4 & 73.8 & div. \\
reversed & 0.042 & 17.2 & 67.2 & 82.5 & 843995.5 & 219.7 & 165382.1 \\
no correction & -- & -- & 67.2 & 30.0 & -- & 219.7 & 46.8 \\
\bottomrule
\end{tabular}%
}
\caption{Correction ordering ablation for $N=12$ on the IDN Reg. checkpoint and No Reg. baseline, evaluated at sequence length 2048.}
\label{tab:h5v3_n12}
\end{table}

\begin{table}[t]
\centering
\scriptsize
\resizebox{\linewidth}{!}{%
\begin{tabular}{lrrrrrrr}
\toprule
& & \multicolumn{3}{c}{IDN Reg.} & \multicolumn{3}{c}{No Reg.} \\
\cmidrule(lr){3-5}\cmidrule(lr){6-8}
Order & AR-ness & Seq-Perm & $K=1$ & $K=4$ & Seq-Perm & $K=1$ & $K=4$ \\
\midrule
forward & 1.000 & \textbf{16.8} & \textbf{18.0} & \textbf{17.0} & \textbf{15.2} & 186.4 & 963.3 \\
shuffle\_7 & 0.254 & 25.6 & 1158.1 & 2958.4 & 64072.0 & 1179.2 & div. \\
shuffle\_15 & 0.225 & 18.7 & 76.5 & div. & 779110.9 & 1044.2 & div. \\
shuffle\_10 & 0.190 & 47.2 & 115.5 & div. & 9214.1 & 173.0 & div. \\
shuffle\_13 & 0.190 & 24.2 & 53.0 & 4823.7 & 206.1 & \textbf{126.7} & div. \\
shuffle\_2 & 0.156 & 19.4 & 86.9 & 75013.0 & 844459.1 & 802.1 & div. \\
shuffle\_14 & 0.131 & 319.6 & div. & 9509.3 & 68188.1 & 198.2 & div. \\
shuffle\_0 & 0.129 & 11585.0 & div. & div. & 73512.0 & 310.8 & div. \\
shuffle\_3 & 0.127 & 26.5 & 1625.1 & div. & 60895.9 & 1257.8 & div. \\
shuffle\_4 & 0.127 & 21.0 & 38.0 & div. & 22761.7 & 400.9 & div. \\
shuffle\_6 & 0.127 & 18.3 & div. & 659.0 & 399655.1 & 711.5 & div. \\
shuffle\_9 & 0.125 & 35.6 & div. & 441.5 & 262632.2 & 1847.5 & div. \\
shuffle\_12 & 0.125 & 17.0 & 60.2 & div. & 71510.7 & 853.7 & div. \\
shuffle\_5 & 0.094 & 21.9 & 30.0 & 24986.9 & 56120.8 & 233.9 & div. \\
shuffle\_8 & 0.094 & 22.0 & div. & 320.6 & 47150.1 & 2951.5 & div. \\
shuffle\_1 & 0.062 & 61.0 & div. & 229855.1 & 20786.3 & 625.6 & div. \\
shuffle\_11 & 0.062 & 33.3 & 978.1 & div. & 12269.6 & 223.4 & div. \\
reversed & 0.031 & 316.4 & div. & div. & div. & div. & 43424.8 \\
no correction & -- & -- & div. & div. & -- & 5382.8 & \textbf{328.6} \\
\bottomrule
\end{tabular}%
}
\caption{Correction ordering ablation for $N=16$ on the IDN Reg. checkpoint and No Reg. baseline, evaluated at sequence length 2048.}
\label{tab:h5v3_n16}
\end{table}

\subsection{Correction Propagation}
\label{app:correction-propagation}

\cref{tab:h7_with_correction} compares subset ablations with and without Newton correction. The identical columns in the no-correction block show that information advances only one layer per iteration, while the correction block shows that all active suffix layers can influence the output immediately.

\begin{table}[t]
\centering
\scriptsize
\resizebox{\linewidth}{!}{%
\begin{tabular}{llrrrrrrrr}
\toprule
Correction & $K$ & last 1 & last 2 & last 3 & last 4 & last 5 & last 6 & last 7 & last 8 \\
\midrule
\multirow{8}{*}{w/ Corr.}
& 1 & 38.70 & 25.64 & 22.32 & 20.57 & 19.44 & 18.71 & 17.89 & 17.19 \\
& 2 & 38.56 & 25.71 & 22.23 & 20.26 & 19.11 & 18.37 & 17.45 & 16.91 \\
& 3 & 38.59 & 25.71 & 22.22 & 20.27 & 19.12 & 18.38 & 17.46 & 16.91 \\
& 4 & 38.59 & 25.71 & 22.22 & 20.27 & 19.12 & 18.38 & 17.46 & 16.91 \\
& 5 & 38.59 & 25.71 & 22.22 & 20.27 & 19.12 & 18.38 & 17.46 & 16.91 \\
& 6 & 38.59 & 25.71 & 22.22 & 20.27 & 19.12 & 18.38 & 17.46 & 16.91 \\
& 7 & 38.59 & 25.71 & 22.22 & 20.27 & 19.12 & 18.38 & 17.46 & 16.91 \\
& 8 & 38.59 & 25.71 & 22.22 & 20.27 & 19.12 & 18.38 & 17.46 & 16.91 \\
\midrule
\multirow{8}{*}{w/o Corr.}
& 1 & \cellcolor{lightblue}37.41 & \cellcolor{lightblue}37.41 & \cellcolor{lightblue}37.41 & \cellcolor{lightblue}37.41 & \cellcolor{lightblue}37.41 & \cellcolor{lightblue}37.41 & \cellcolor{lightblue}37.41 & \cellcolor{lightblue}37.41 \\
& 2 & 32.07 & \cellcolor{lightblue}22.97 & \cellcolor{lightblue}22.97 & \cellcolor{lightblue}22.97 & \cellcolor{lightblue}22.97 & \cellcolor{lightblue}22.97 & \cellcolor{lightblue}22.97 & \cellcolor{lightblue}22.97 \\
& 3 & 37.58 & 25.21 & \cellcolor{lightblue}21.90 & \cellcolor{lightblue}21.90 & \cellcolor{lightblue}21.90 & \cellcolor{lightblue}21.90 & \cellcolor{lightblue}21.90 & \cellcolor{lightblue}21.90 \\
& 4 & 32.98 & 23.40 & 21.01 & \cellcolor{lightblue}19.73 & \cellcolor{lightblue}19.73 & \cellcolor{lightblue}19.73 & \cellcolor{lightblue}19.73 & \cellcolor{lightblue}19.73 \\
& 5 & 37.74 & 25.27 & 21.95 & 19.86 & \cellcolor{lightblue}18.88 & \cellcolor{lightblue}18.88 & \cellcolor{lightblue}18.88 & \cellcolor{lightblue}18.88 \\
& 6 & 37.83 & 25.31 & 21.97 & 19.87 & 18.89 & \cellcolor{lightblue}18.17 & \cellcolor{lightblue}18.17 & \cellcolor{lightblue}18.17 \\
& 7 & 49.44 & 28.55 & 25.01 & 21.06 & 19.69 & 18.75 & \cellcolor{lightblue}17.72 & \cellcolor{lightblue}17.72 \\
& 8 & 38.59 & 25.71 & 22.22 & 20.27 & 19.12 & 18.38 & 17.46 & \cellcolor{lightblue}16.91 \\
\bottomrule
\end{tabular}%
}
\caption{Subset ablation with and without Newton correction for the 0.5B IDN Reg. model at sequence length 2048. Columns activate progressively more suffix layers ending at layer 31. With correction, all active layers affect PPL at $K=1$. Without correction, influence propagates one layer per iteration: at $K=1$ all columns are identical, at $K=2$ columns last 2--last 8 are identical, and so on.}
\label{tab:h7_with_correction}
\label{tab:h7_without_correction}
\end{table}

% \newpage
% \input{checklist.tex}

\end{document}